\documentclass[final]{cvpr}

\usepackage{times}
\usepackage{epsfig}
\usepackage{graphicx}
\usepackage{amsmath}
\usepackage{amssymb}

\usepackage{xspace}
\usepackage{comment}

\usepackage{makecell}

\usepackage{booktabs}       % professional-quality tables
\usepackage{nicefrac}       % compact symbols for 1/2, etc.
\usepackage{microtype}      % microtypography
\usepackage{subfigure}
\usepackage[flushleft]{threeparttable}
\usepackage{multirow}
\usepackage{xcolor}
\usepackage{amsmath} % for equation
\usepackage{bbm}
\usepackage{algorithm}
\usepackage{algorithmic}
\usepackage{amsfonts,amssymb}
\usepackage{booktabs,tabulary}
\usepackage{wrapfig}
\usepackage{multirow}

\def\cyf{\textcolor{black}}
\usepackage[pagebackref=true,breaklinks=true,colorlinks,bookmarks=false]{hyperref}

\def\mA{{\mathcal A}}
\def\mB{{\mathcal B}}
\def\mC{{\mathcal C}}
\def\mD{{\mathcal D}}
\def\mE{{\mathcal E}}
\def\mF{{\mathcal F}}
\def\mG{{\mathcal G}}
\def\mH{{\mathcal H}}

\def\mL{{\mathcal L}}
\def\mM{{\mathcal M}}

\def\mR{{\mathcal R}}
\def\mS{{\mathcal S}}

\def\mU{{\mathcal U}}

\def\mX{{\mathcal X}}

\def\0{{\bf 0}}
\def\1{{\bf 1}}

\def\bA{{\bf A}}

\def\bW{{\bf W}}
\def\bX{{\bf X}}

\def\bZ{{\bf{Z}}}

\def\mmE{{\mathbb E}}

\def\mmR{{\mathbb R}}

\def\bX{{\bf X}}

\def\bW{{\bf W}}

\newtheorem{thm}{Theorem}
\newtheorem{prop}{Proposition}

\def\eg{\emph{e.g.}} 
\def\ie{\emph{i.e.}} 
 
 \def\vs{\emph{vs.}}
\def\wrt{{w.r.t.}} 
\def\etal{{\em et al.\/}\, }

\def\cvprPaperID{11671} % *** Enter the CVPR Paper ID here
\def\confYear{CVPR 2021}

\newcommand{\sexyname}{NAC\xspace}
\newcommand{\sexynamenas}{CTNAS\xspace}

\newcommand{\topline}{\toprule [0.1em]}
\newcommand{\midline}{\midrule [0.05em]}
\newcommand{\bottomline}{\bottomrule [0.1em]}

\begin{document}

%%%%%%%%% TITLE
\def\mytitle{Contrastive Neural Architecture Search with Neural Architecture Comparators}
\title{\mytitle}

\author{
    Yaofo Chen$^{1,2}$\thanks{Authors contributed equally.}~, Yong Guo$^{1*}$, Qi Chen$^{1}$, Minli Li$^{1}$, Wei Zeng$^{3}$, Yaowei Wang$^{2\dagger}$, Mingkui Tan$^{1,4}$\thanks{Corresponding author.} \\
    $^{1}$South China University of Technology, $^{2}$Peng Cheng Laboratory, $^{3}$Peking University, \\ 
    $^{4}$Key Laboratory of Big Data and Intelligent Robot, Ministry of Education \\
    {\tt\small \{sechenyaofo, guo.yong, sechenqi, seminli\_li\}@mail.scut.edu.cn, } \\
    {\tt\small weizeng@pku.edu.cn, wangyw@pcl.ac.cn, mingkuitan@scut.edu.cn }
}

\maketitle

\begin{abstract}
One of the key steps in Neural Architecture Search (NAS) is to estimate the performance of candidate architectures.
Existing methods either directly use the validation performance or learn a predictor to estimate the performance.
However, these methods can be either computationally expensive or very inaccurate, which may severely affect the search efficiency and performance.
Moreover, as it is very difficult to annotate architectures with accurate performance on specific tasks, learning a promising performance predictor is often non-trivial due to the lack of labeled data.
In this paper, we argue that it may not be necessary to estimate the absolute performance for NAS.
On the contrary, we may need only to understand whether an architecture is better than a baseline one. However, how to exploit this comparison information as the reward and how to well use the limited labeled data remains two great challenges. In this paper, we propose a novel Contrastive Neural Architecture Search (\sexynamenas) method which performs architecture search by taking the comparison results between architectures as the reward. Specifically, we design and learn a Neural Architecture Comparator (\sexyname) to compute the probability of candidate architectures being better than a baseline one.
Moreover,
we present a baseline updating scheme to improve the baseline iteratively in a curriculum learning manner.
More critically, we theoretically show that learning \sexyname is equivalent to optimizing the ranking over architectures.
Extensive experiments in three search spaces demonstrate the superiority of our \sexynamenas over existing methods.
\end{abstract}

\section{Introduction}\label{introduction}

Deep neural networks (DNNs) have made significant progress in various challenging tasks, including image classification~\cite{guo2020multi, DBLP:conf/cvpr/SandlerHZZC18, szegedy2015going}, face recognition~\cite{SphereFace,CosFace}, and many other areas~\cite{deeplabv3plus,chen2020RSPNet,zeng2019graph,zeng2020dense}.
One of the key factors behind the progress lies in the innovation of effective neural architectures, such as ResNet~\cite{resnet} and MobileNet~\cite{howard2017mobilenets}.
However, designing effective architectures is often labor-intensive and relies heavily on human expertise.
Besides designing architectures manually, Neural Architecture Search (NAS) seeks to design architectures automatically and outperforms the hand-crafted architectures in various tasks~\cite{baker2016designing,pham2018efficient}.

Existing NAS methods seek to find the optimal architecture by maximizing the expectation of the performance of the sampled architectures.
Thus, how to estimate the performance of architectures is a key step in NAS.
In practice, the searched architectures can be evaluated by the absolute performance provided by a supernet~\cite{Cai2020Once,chu2019fairnas,pham2018efficient} or a predictor~\cite{DBLP:conf/eccv/LiuZNSHLFYHM18,luo2018neural}.
However, using the absolute performance as the training signal may suffer from two limitations.

\begin{figure}[t]
\centering
\includegraphics[width=0.9\linewidth]{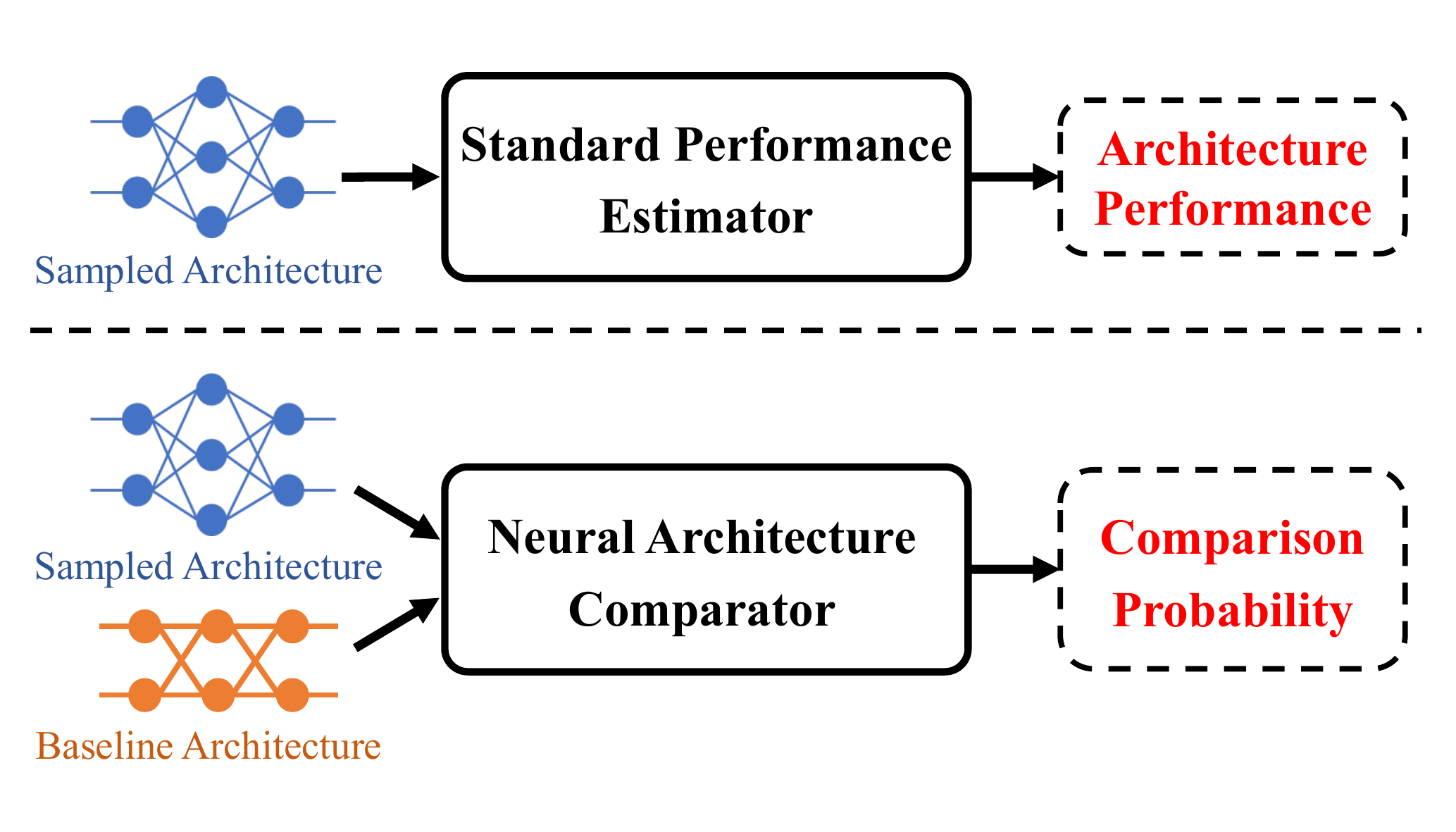}
\caption{
    Comparison between the standard performance estimator (top) and our \sexyname (bottom).
    Unlike the standard estimator that predicts the absolute performance, our \sexyname takes two architectures as inputs and outputs the comparison probability of the sampled architectures being better than the baseline architecture.
}
\label{fig:illustration_nac}
\end{figure}

First, it is non-trivial to obtain stable and accurate absolute performance for all the candidate architectures.
In practice, the performance of architectures may fluctuate a lot under the training with different random seeds~\cite{DBLP:conf/uai/LiT19,liu2018darts}. Thus, there would be a large performance deviation if we evaluate the architecture only with a single value \wrt~absolute performance.
As a result, using the absolute performance as the training signals may greatly hamper the search performance.
Based on such signals, a randomly searched architecture may even outperform the architectures obtained by existing NAS methods~\cite{DBLP:conf/uai/LiT19,yu2020evaluating} in practice.
Thus, how to obtain stable and accurate training signals to guide the search is an important problem.

Second, it is time-consuming to obtain the absolute performance from the supernet.
Specifically, one can evaluate an architecture 
by feeding in the validation data on a specific task to obtain the accuracy.
However, given a large number of validation data, obtaining the validation accuracy for candidate architectures via forward propagation can be computationally expensive.
To address this issue, one can learn a regression model to predict the performance of architectures~\cite{DBLP:conf/eccv/LiuZNSHLFYHM18,luo2018neural}. 
However, the training of predictor models still requires plenty of architectures with the ground-truth performance as the training architecture data, which are very expensive to obtain in practice.
Thus, how to efficiently evaluate architectures with limited architectures with ground-truth performance becomes an important problem.

In this paper, we propose a Contrastive Neural Architecture Search (\sexynamenas) method that searches by architecture comparisons.
To address the first limitation, we devise a Neural Architecture Comparator (\sexyname) to perform pairwise architecture comparisons.
Unlike existing methods that rely on the absolute performance, we use the comparison results between the searched architectures and a baseline one as the reward (See Figure~\ref{fig:illustration_nac}).
In practice, the pairwise comparison results are easier to obtain and more stable than the absolute performance (See analysis in Section~\ref{sec:motivation}).
To constantly find better architectures,
we propose to improve the baseline gradually via a curriculum learning manner.
To address the second limitation, the proposed \sexyname evaluates architectures via pairwise comparisons and avoid performing forward propagation on task data. Thus, the evaluation can be much more efficient and greatly accelerate the search process (See Table~\ref{exp:imagenet}).
Moreover, we also propose a data exploration method that exploits the architectures without ground-truth performance to improve the generalization ability of \sexyname to unseen architectures. In this way, we are able to effectively reduce the requirement of the training data for \sexyname.

Our contributions are summarized as follows.
\begin{itemize}
    \item We propose a Contrastive Neural Architecture Search (\sexynamenas) method that searches for promising architectures by taking the comparison results between architectures as the reward.
    \item To guarantee that \sexynamenas can constantly find better architectures, we propose a curriculum updating scheme to gradually improve the baseline architecture. In this way, \sexynamenas has a more stable search process and thus greatly improves the search performance.
    \item Extensive experiments on three search spaces demonstrate that the searched architectures of our \sexynamenas outperform the architectures searched/designed by state-of-the-art methods.
\end{itemize}

\section{Related Work}
\textbf{Neural Architecture Search}.
NAS seeks to automatically design neural architectures in some search space.
The pioneering work~\cite{zoph2016neural} exploits the paradigms of reinforcement learning (RL) to solve it.
RL-based methods~\cite{baker2016designing,DBLP:conf/cvpr/GhiasiLL19,pham2018efficient,tan2019mnasnet,zoph2016neural,zoph2018learning,guo2020hierarchical} seek to learn a controller with a policy $\pi(\alpha; \theta)$ to generate architectures, where $\alpha$ denotes a sampled architecture, $\theta$ denotes the parameters of the policy.
Specifically, they learn the controller by maximizing the expectation of some performance metric $\mR(\alpha, w_\alpha)$,
\begin{equation} \label{eq:nas-general}
	\begin{aligned}
	\max_{\theta}~\mmE_{\alpha\sim \pi(\alpha;\theta)}\mR\left(\alpha, w_{\alpha}\right),
	\end{aligned}
\end{equation}
where $w_{\alpha}\small{=}\arg\min_{w}\mL\left(\alpha, w\right)$ and $\mL\left(\alpha, w\right)$ is the training loss.
Moreover, some studies~\cite{liu2017hierarchical,DBLP:journals/corr/abs-1910-06961,real2018regularized} search for promising architectures using evolutionary algorithms.
Different from the above methods, gradient based methods~\cite{chen2019progressive,liu2018darts,DBLP:conf/cvpr/WuDZWSWTVJK19,xu2020pcdarts} represent architectures by continuous relaxation and optimize by gradient descent.
Unlike existing methods that obtain/predict the absolute performance, we seek to conduct comparisons between architectures to guide the search process.
Based on the RL algorithm, our method maximizes the expectation of the comparison probability of the sampled architectures being better than a baseline one instead of the absolute performance.

\begin{figure*}[t]
    \centering
    \subfigure[Overall scheme of Contrastive Neural Architecture Search]{
        \includegraphics[width=0.48\linewidth]{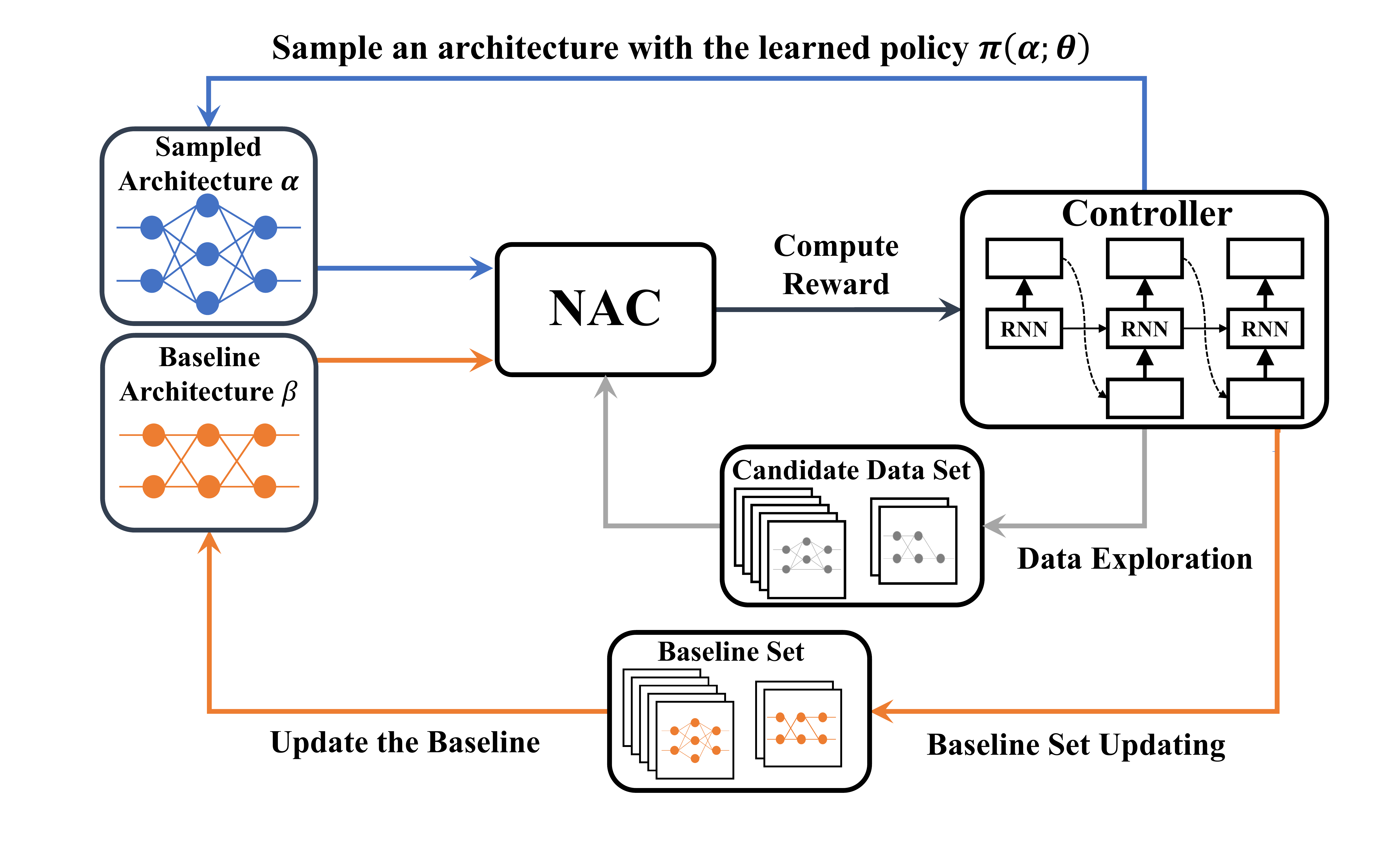}
    }
    \subfigure[Neural Architecture Comparator]{
        \includegraphics[width=0.48\linewidth]{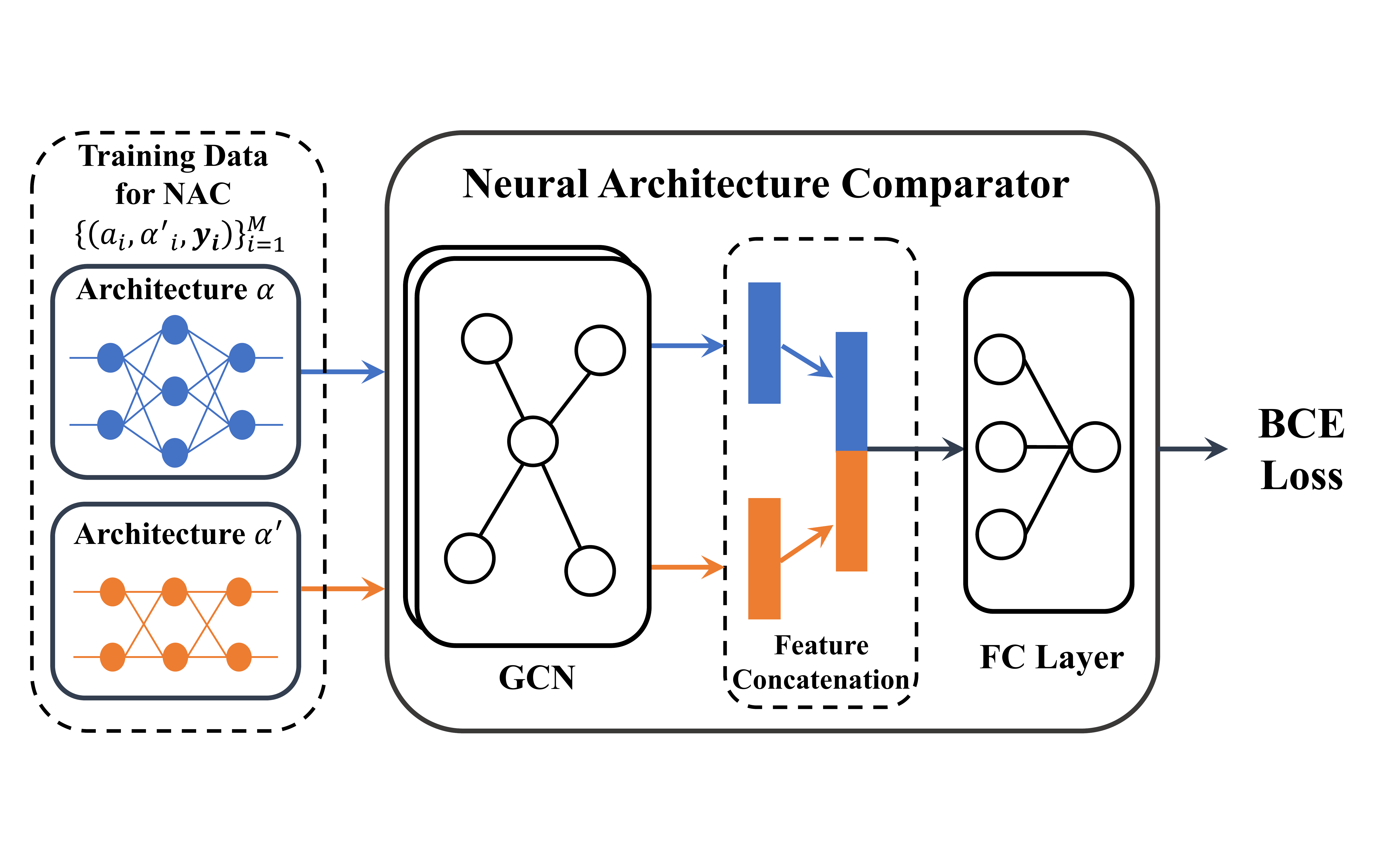}
    }
    \caption{
    The overview of \sexynamenas and \sexyname.
    (a) The proposed \sexyname first takes the sampled architecture and the baseline one as inputs, and outputs the comparison probability of them.
    Then, our \sexynamenas adopts the probability as the reward to train the controller.
    During the training, we update the baseline sampled from the controller.
    Besides, we perform data exploration on the sampled architectures to construct a candidate data set.
    (b) We optimize \sexyname with the binary cross-entropy (BCE) loss computed by the comparison probability and the label indicated which one is better between two input architectures.}
    \label{fig:NACNAS_scheme}
\end{figure*}

\textbf{Contrastive Learning.}
Contrastive learning aims to learn the similarity/dissimilarity over the samples by performing comparisons among them.
Specifically, Hadsell~\etal \cite{hadsell2006dimensionality} propose a contrastive loss to solve the dimensionality reduction problem by performing pairwise comparison among different samples.
Based on the contrastive loss, Sohn~\cite{sohn2016improved} 
proposes a Multi-class N-pair loss to allow joint comparison among multiple negative data pairs.
As for NAS, finding the optimal architecture can be considered as a ranking problem over a set of candidate architectures~\cite{yang2019evaluation}.
In this sense, it is possible to solve the ranking problem of NAS by conducting comparisons among architectures.

\textbf{Comparisons with ReNAS~\cite{yi2019renas}.} ReNAS exploits a ranking loss to learn a predictor model that predicts the relative score of architectures.
However, ReNAS is essentially different from our method and has several limitations.
\textbf{First}, ReNAS searches with the predicted scores of architectures;
while our \sexynamenas proposes a contrastive architecture search scheme that searches by comparing architectures.
\textbf{Second}, ReNAS heavily relies on the training data and thus may be hard to generalize to unseen architectures.
In contrast, the proposed \sexynamenas introduces a data exploration method to improve the generalization (See results in Table~\ref{tab:nasbench_search}).

\section{Proposed Method}

\noindent \textbf{Notation.}
Throughout the paper, we use the following notations.
We use calligraphic letters (\textit{e.g.}, $ \mA $) to denote a set.
We use the lower case of Greek characters (\textit{e.g.}, $ \alpha $) to denote architectures.
Let $\Omega$ be the search space.
For any architecture $ \alpha \in \Omega$, let $w_\alpha$ be its optimal model parameters trained on some data set.
Let $\mathbbm{1}\{\cdot\}$ be an indicator function, where $\mathbbm{1}\{A\} = 1$ if $A$ is true and  $\mathbbm{1}\{A\} = 0$ if $A$ is false.
Let $\Pr[\cdot]$ be the probability for some event.

In this paper, we propose a Contrastive Neural Architecture Search (\sexynamenas) that conducts architecture search by taking the comparisons between architectures as the reward.
To ensure that \sexynamenas is able to constantly find better architectures, we seek to gradually improve/update the baseline via a curriculum learning manner.
We show the training method of \sexynamenas in Algorithm~\ref{alg:training}.

% \vspace{3 pt}

\begin{algorithm}[t]
\small
\caption{The overall algorithm for \sexynamenas.}
\label{alg:training}
    \begin{algorithmic}[1]
    	\REQUIRE
    	Learning rate $\eta$ for policy gradient, parameters $M$, $N$ and $K$ ($K\ll N$). \\
    	\STATE Randomly sample a set of architectures from $\Omega$ and obtain their accuracy $\{\alpha_i, \mR(\alpha_i, w_{\alpha_i})\}_{i=1}^M$ by training a supernet.
    	\STATE Construct training data $\mA\small{=}\{(\alpha_i, \alpha'_i,  y_i)\}_{i=1}^{M(M-1)/2}$ for \sexyname by traversing all pairwaise combinations.
        \STATE Initialize parameters $\theta$ for $\pi(\cdot; \theta)$ and  $\varpi$ for \sexyname. \\
        \STATE Initialize the baseline architecture $\beta \sim \pi(\cdot; \theta)$.
        \STATE Let $\mC \small{=} \mA$, $\mD \small{=} \emptyset$,  $\mB \small{=} \emptyset$.
        \FOR{$t$ = $1,\dots,T$}
        \STATE Train \sexyname with data $\mC\small{=}\{(\alpha_i, \alpha'_i,  y_i)\}_{i=1}^{|\mC|}$. 
        \STATE // \emph{Train the controller with \sexyname}
        \STATE Sample $N$ architectures $\{\alpha_j\}_{j=1}^N$ by $ \alpha \sim \pi(\cdot; \theta)$.
    	\STATE Update $\theta$ using policy gradient:\\
    	$\theta {\leftarrow} \theta {+} \eta \frac{1}{N}\sum_{j=1}^N \left[ \nabla_{\theta} \log \pi(\alpha_j; \theta) \text{\sexyname}(\alpha_j, \beta; \varpi) \right]$.
        \STATE // \emph{Explore  more data for training \sexyname}
        \STATE Sample $N$ architectures $\mS{=}\{\alpha_i\}_{i=1}^N \sim \pi(\cdot; \theta)$.
        \STATE Construct $\mD$ with $\mS$ by data exploration using Alg.~\ref{alg:data_exploration}.
        \STATE Let $\mC = \mC \cup \mD$ and $\mB = \mB \cup\{\beta\}$.% // \emph{Augment the data}    
        \STATE {Update the baseline $\beta$ with $\mB$ and $\mS$ using Alg.~\ref{alg:updating_baseline}.}
        \ENDFOR
    \end{algorithmic}
\end{algorithm}

\subsection{Motivation}\label{sec:motivation}

In neural architecture search (NAS), one of the key steps is to evaluate the candidate architectures.
Most existing methods evaluate architectures through the absolute performance $\mR(\alpha, w_\alpha)$ with $w_\alpha$ from a learned supernet~\cite{Cai2020Once, chu2019fairnas,DBLP:journals/corr/abs-1904-00420,pham2018efficient}.
However, finding promising architectures with the absolute performance comes with two challenges.
First, there would be deviation on the absolute performance with different training random seeds.
Searching with this fluctuated performance may incur training difficulties for the NAS model.
Second, obtaining absolute performance with a learned supernet is computationally expensive since it needs to compute the accuracy on plenty of validation data.
Given a large number of candidate architectures in the search process, it would be very time-consuming (\eg, several GPU days).

In this paper, we seek to improve the search performance by designing a more effective evaluation method.
To begin with, let us revisit the definition of optimal architecture.
Suppose that $\alpha^*$ is the optimal architecture in the search space $\Omega$, 
we would have
$\mR(\alpha^*, w_{\alpha^*}) \small{\geq} \mR(\alpha, w_{\alpha}), \forall~\alpha \in \Omega.$
However, since $\mR(\alpha, w_{\alpha})$ may not be very accurate, it is more reasonable and rigorous to require that $\mR(\alpha^*, w_{\alpha^*}) \small{\geq} \mR(\alpha, w_{\alpha})$ holds in high probability. 
For example, it is often much easier to recognize which one is better among two architectures compared to estimating the absolute performance of them.
In this way, the comparison results become more stable and may reduce the influence of training fluctuations.
Thus, to ensure the optimality, we only need to compute
\begin{equation}
    \mathrm{Pr}[\mR(\alpha^*, w_{\alpha^*}) \geq \mR(\alpha, w_{\alpha}) ].
\end{equation}
The above probability implies that we may not need to obtain the absolute performance to solve the NAS problem.

\subsection{Contrastive Neural Architecture Search}\label{sec:overall_scheme}

In this paper, we propose a Contrastive Neural Architecture Search (\sexynamenas) method that finds promising architectures via architecture comparisons.
Unlike existing methods that rely on the absolute performance, we seek to obtain the ranking of candidate architectures using a series of pairwise comparisons.
Specifically, we can learn a comparison mapping, called Neural Architecture Comparator (\sexyname), to compare any two architectures $\alpha, \alpha' \in \Omega$ and output the probability of $\alpha$ being better than $\alpha'$:
\begin{equation}\label{eq:NAC}
     p = \mathrm{Pr}[\mR(\alpha, w_{\alpha}) \geq \mR(\alpha', w_{\alpha'})] = \text{\sexyname}(\alpha, \alpha'; \varpi),
\end{equation}
where $\varpi$ is the parameter of \sexyname.
The comparison probability $p$ is more stable than the absolute performance since it may reduce the negative impact of accuracy deviation.
For simplicity, we leave the details of \sexyname in the following.

From Eqn.~(\ref{eq:NAC}), it is possible to use the comparison probability predicted by \sexyname as the reward signal to train the NAS model.
Formally, given a baseline architecture $\beta \in \Omega$, we hope to learn a policy $\pi(\alpha;\theta)$ by solving the following optimization problem:
\begin{equation}\label{eq:compared_nas_optimization}
    \begin{aligned}
    \max_{\theta}~&\mmE_{\alpha\sim \pi(\alpha;\theta)}\Pr[\mR(\alpha, w_{\alpha}) \geq \mR(\beta,  w_{\beta}) ],
    \end{aligned}
\end{equation}
where $\theta$ denotes the parameters of the policy.
To address the above optimization problem, following~\cite{pham2018efficient,zoph2016neural}, we train a controller with policy gradient~\cite{williams1992simple}.
Unlike existing reinforcement learning based NAS methods,
we adopt the comparison probability 
$p\small{=}\text{\sexyname}(\alpha, \beta; \varpi)$
as the reward. Given a specific $\beta$, the controller seeks to conduct a comparison with it to search for better architectures.

However, solving Problem (\ref{eq:compared_nas_optimization}) can only enable the model to find architectures that are better than the baseline $\beta$.
In other words, it may not find the optimal architecture.
Moreover, in Problem (\ref{eq:compared_nas_optimization}), if $\beta$ is too weak or strong, the above optimization problem becomes meaningless (\ie, the optimal objective value will be trivially 1 or 0, respectively).
To address this issue, we propose a baseline updating scheme to improve/update the baseline gradually.
In this way, our \sexynamenas is able to find better architectures iteratively.
We will detail the baseline updating scheme in the following.

\begin{algorithm}[t]
\small
	\caption{Baseline updating via curriculum learning.}
	\label{alg:updating_baseline}
	\begin{algorithmic}[1]
        \REQUIRE Existing baseline architectures $\mathcal{B}$, sampled architectures $\mathcal{S}$, and learned architecture comparator $\mathrm{\sexyname}(\cdot,\cdot;\varpi)$.
        % baseline architecture $\beta$.
        \STATE Initialize comparison score $\hat{s} = 0$.
        \STATE Construct a candidate baseline set $\mathcal{H}=\mathcal{B} \cup \mathcal{S} = \{\alpha_i\}_{i=1}^{|\mathcal{H}|}$.
        \FOR{$i=1,\dots, |\mathcal{H}|$}
            \STATE   {Compute score for architecture $\alpha_i \in \mathcal{H}$} by
             $$s_i=\frac{1}{|\mathcal{H}|-1}\sum_{1\leq j \leq |\mathcal{H}|, i \neq j}\text{\sexyname}(\alpha_i, \alpha_j;\varpi).$$ 
            % \STATE \\ \emph{}
            \STATE \textbf{if} {$s_i \geq \hat{s}$} \textbf{then} $\hat{s} = s_i$ and $\beta = \alpha_i$. \textbf{end if}
                % \STATE Let $\hat{s} = s_i$ and $\beta = \alpha_i$. 
            % \ENDIF
        \ENDFOR
        \STATE Return $\beta$.
    \end{algorithmic}
\end{algorithm}

\subsection{Baseline Updating via Curriculum Learning}\label{sec:baseline_update}

Since \sexynamenas takes the comparison result with the baseline architecture as the reward,
the search performance heavily relies on the baseline architecture.
Given a fixed baseline architecture, the controller is only able to find better architectures than the baseline.
If the baseline is not good enough, the searched architecture cannot be guaranteed to be a promising one.
Thus, it becomes necessary to gradually improve/update the baseline architecture during the search.

To this end, we propose a curriculum updating scheme to improve the baseline during the search process (See Algorithm~\ref{alg:updating_baseline}).
The key idea follows the basic concept of curriculum learning that humans and animals can learn much better when they gradually learn new knowledge~\cite{bengio2009curriculum,gulccehre2016knowledge}.
\cyf{Specifically, the difficulty of finding architectures than a gradually improved baseline architecture would increase.}
% Specifically, we break the whole NAS problem into a series of simpler sub-problems, \ie, solving Problem~(\ref{eq:compared_nas_optimization}) with a gradually improved baseline architecture.
As we gradually improve the baseline, our \sexynamenas is able to constantly find better architectures during the search.

To improve the baseline architecture $\beta$, we seek to select the best architecture from the previously searched architectures (See Algorithm~\ref{alg:updating_baseline}). 
Specifically, we build a candidate baseline set $\mH$ and dynamically incorporate sampled architectures into it. To avoid the negative impact from the possible error of \sexyname in a single comparison, for any architecture $\alpha_i \in \mH$, we compute the average comparison probability $s_i$ by comparing $\alpha_i$ with other architectures in $\mH$:
\begin{equation}
    s_i=\frac{1}{|\mathcal{H}|-1}\sum_{1\leq j \leq |\mathcal{H}|, i \neq j}\text{\sexyname}(\alpha_i, \alpha_j;\varpi).
\end{equation}
Based on the best architecture in the past as the baseline, our \sexynamenas is able to further improve the search performance by finding better architectures than the baseline.

Compared with existing NAS methods, our \sexynamenas is more stable and reliable since the contrastive search scheme searches for better architectures than the best one of the previously searched architectures. Thus, our \sexynamenas consistently finds better architectures than existing methods on different search spaces (See results in Tables~\ref{tab:nasbench_search} and~\ref{tab:imagenet}).

\section{Neural Architecture Comparator}

To provide a valid reward for \sexynamenas, we propose a neural architecture comparator (\sexyname) that compares any two architectures.
To guarantee that \sexyname can handle any architecture, we represent an architecture as a directed acyclic graph (DAG)~\cite{pham2018efficient} and build \sexyname using a graph convolutional network (GCN) to calculate the comparison probability. 
Moreover, we develop a data exploration method that trains \sexyname in a semi-supervised way to reduce the requirement of the training data.

\begin{figure}[t]
\centering
\includegraphics[width=\linewidth]{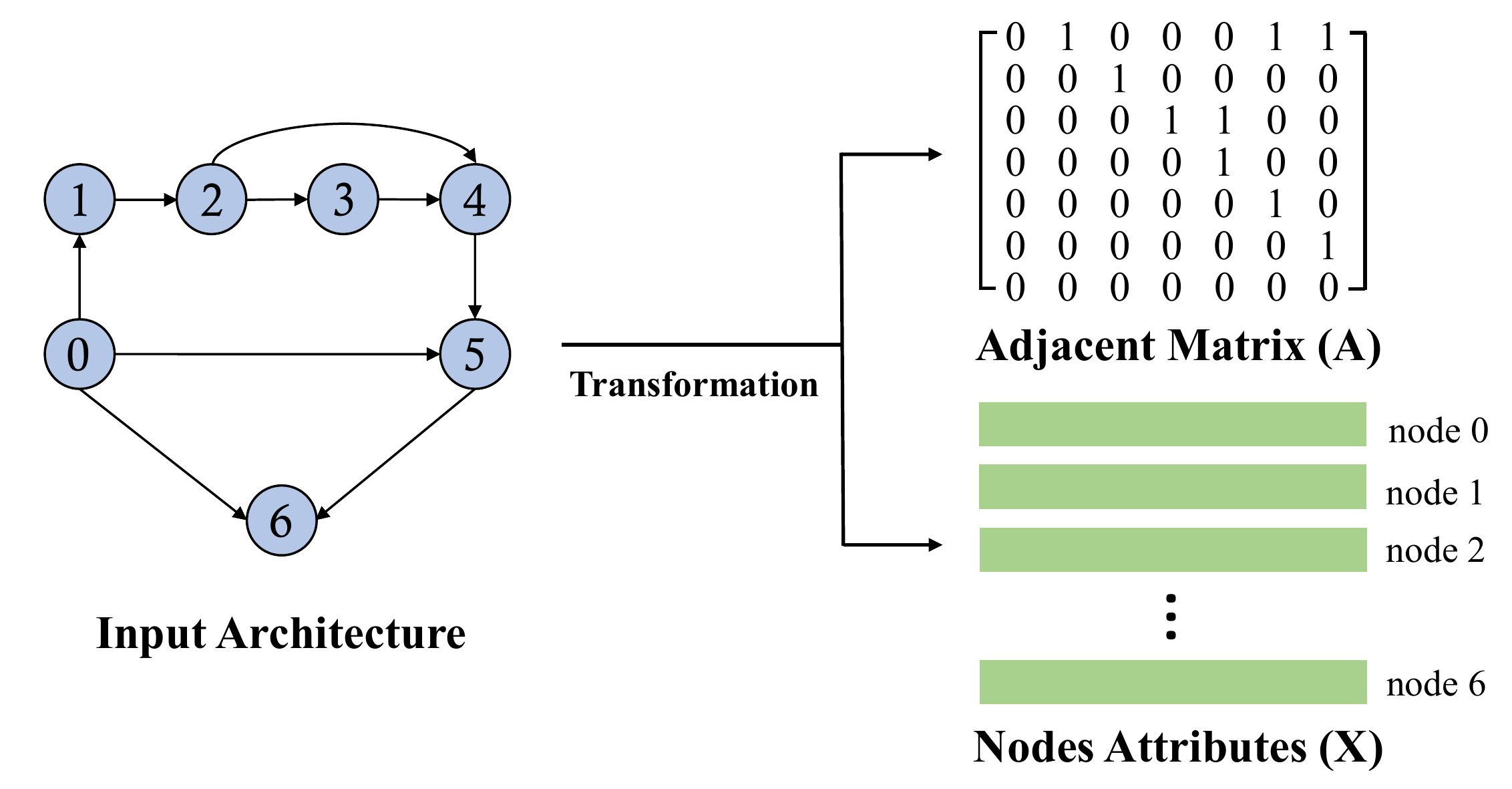}
\caption{
    Architecture representation method of the proposed \sexynamenas.
    The nodes of the architecture DAG indicate the operations.
    The edges represent the flow of information.
}
\label{fig:arch_representation}
\end{figure}

\subsection{Architecture Comparison by GCN}\label{sec:NAC_GCN}

Given an architecture space $\Omega$, we can represent an architecture $\alpha \in \Omega$ as a directed acyclic graph (DAG)~\cite{pham2018efficient}. As shown in Figure~\ref{fig:arch_representation}, each node represents a computational operation (\eg, convolution or max pooling) and each edge represents the flow of information. Following~\cite{kipf2016semi}, we represent the graph $\alpha$ using a data pair $(\bA_{\alpha}, \bX_{\alpha})$, where $\bA_{\alpha}$ denotes the adjacency matrix of the graph and $\bX_{\alpha}$ \cyf{denotes the learnable embeddings of nodes/operations in $\alpha$.}

The proposed \sexyname compares any two architectures and predicts the comparison probability as the reward. To exploit the connectivity information inside the architecture graph, we build the \sexyname model with a graph convolutional network. Specifically, given two architectures $\alpha$ and $\alpha'$ as inputs,
the proposed \sexyname predicts the probability of $\alpha$ being better than $\alpha'$ (See Figure~\ref{fig:NACNAS_scheme} (b)).
To calculate the comparison probability, we concatenate the features of $\alpha$ and $\alpha'$ and send them to a fully-connected (FC) layer.
Then, the sigmoid function $\sigma(\cdot)$ takes the output of the FC layer as input and outputs the comparison probability:
\begin{equation}\label{eq:nac_head}
p = \text{\sexyname}(\alpha, \alpha'; \varpi) = \sigma\left([\bZ_{\alpha}; \bZ_{\alpha'}]{\bW}^{FC}\right),
\end{equation}
where $Z_{\alpha}$ denotes the features extract from the architecture $\alpha$, ${\bW}^{FC}$ denotes the weight of the FC layer, $[\bZ_{\alpha}; \bZ_{\alpha'}]$ refers to the concatenation of the features of $\alpha$ and $\alpha'$.
Based on the graph data pair $(\bA_{\alpha}, \bX_{\alpha})$, we use a two-layer GCN to extract the architecture features ${\bZ_{\alpha}}$ following~\cite{guo2019nat}:
\begin{equation}\label{eq:gcn}
{\bZ_{\alpha}} = f({\bX_{\alpha}}, {\bA_{\alpha}}) = {\bA_{\alpha}} {\rm \phi}\left({\bA_{\alpha}} {\bX_{\alpha}} {\bW}^{(0)}\right) {\bW}^{(1)} ,
\end{equation}
where $\bW^{(0)}$ and $\bW^{(1)}$ denote the weights of two graph convolutional layers, $\phi$ is the a non-linear activation function (\eg, the Rectified  Linear Unit  (ReLU)~\cite{nair2010rectified}), and $\bZ_{\alpha}$ refers to the extracted features.

\begin{algorithm}[t]
    \small
	\caption{Data exploration for \sexyname.}
	\label{alg:data_exploration}
	\begin{algorithmic}[1]
	    \REQUIRE Sampled architecture set $\mS\small{=}\{\alpha_i\}_{i=1}^
	    N$, learned architecture comparator $\mathrm{\sexyname}(\cdot,\cdot;\varpi)$, and parameter $K$.\\
		\STATE Initialize the set of predicted label data
		$\mE\small{=}\emptyset$ and the confidence score set $\mF{=}\emptyset$.
        \STATE Build $\mG \small{=} \{(\alpha_k, \alpha'_k)\}, \alpha_k, \alpha'_k \in \mS$, $\alpha_k \neq \alpha'_k$.
        \FOR{$k=1,\dots,|\mG|$}
        \STATE $p_{k} \small{=} \text{\sexyname}(\alpha_k, \alpha'_k; \varpi)$. // \emph{Compute comparison probability}
        \STATE $y'_k\small{=}\mathbbm{1}\{p_k \small{\geq} 0.5\}$.// \emph{Compute label according to probability}
        \STATE Let $\mE = \mE \cup \{(\alpha_k, \alpha'_k, y'_k)\}$.
        \STATE $f_{k} \small{=}\left| p_{k}  \small{-} 0.5 \right|$. // \emph{Compute confidence score}
        \STATE Let $\mF = \mF \cup \{f_k\}$.
        \ENDFOR
        \STATE Select top-\textit{K} architecture pairs according to confidence score $\mD$ = top-\textit{K}($\mE, \mF, K$).
        \STATE Return $\mD$.
	\end{algorithmic}
\end{algorithm}

To train the proposed \sexyname, we need a data set that consists of architectures and their performance evaluated on some data set (\eg, CIFAR-10).
The data pairs of architecture and its performance can be obtained by training a supernet or training a set of architectures from scratch.
Based on the data set, we define the label for any pair $(\alpha_i, \alpha'_i)$, $\alpha_i \small{\neq} \alpha'_i$, sampled from $\Omega$, by $y_i\small{=}\mathbbm{1}\{\mR(\alpha_i, w_{\alpha_i}) \small{-} \mR(\alpha'_i, w_{\alpha'_i}) \small{\geq} 0\}$, and construct the training data 
$(\alpha_i, \alpha'_i,  y_i)$.
% $\mA\small{=}\{(\alpha_i, \alpha'_i,  y_i)\}_{i=1}^M$.
Thus, the training of \sexyname can be considered a binary classification problem (\ie, the label $y \in \{0,1\}$). We solve the problem by optimizing the binary cross-entropy loss between the probability $p$ predicted by \sexyname and the ground truth label $y$.

\begin{table*}[t]
	\caption{Comparisons with existing methods in NAS-Bench-101 search space.
	``--'' represents unavailable results.
	\cyf{``Best Rank'' denotes the percentile rank of the best searched architecture among all the architectures in the search space.}
	``\#Queries'' denotes the number of pairs of architecture and its validation accuracy queried from the NAS-Bench-101 dataset.
	All methods are run 10 times.}
    % We run all our methods for 10 times and report the average accuracy as well as the best accuracy.
	\centering
    {
    \resizebox{0.72\linewidth}{!}{
	\begin{tabular}{cccccc}
		\topline
		Method & KTau & Average Accuracy (\%) & Best Accuracy (\%)  & Best Rank (\%)  & \#Queries \\
		\midline
		Random & -- & 89.31 $\pm$ 3.92 & 93.46 & 1.29 & 423 \\
        DARTS~\cite{liu2018darts} & -- & 92.21 $\pm$ 0.61 & 93.02 & 13.47 & --\\
        % NAONet~\cite{luo2018neural} & -- & 92.59 $\pm$ 0.59 & 93.33 & 4.62 & --\\
        ENAS~\cite{pham2018efficient} & -- & 91.83 $\pm$ 0.42 & 92.54 & 22.88 & --\\
        FBNet~\cite{DBLP:conf/cvpr/WuDZWSWTVJK19} & -- & 92.29 $\pm$ 1.25 & 93.98 & 0.05 & -- \\
        SPOS~\cite{DBLP:journals/corr/abs-1904-00420} & 0.195 & 89.85 $\pm$ 3.80 & 93.84 & 0.07 & --\\
        FairNAS~\cite{chu2019fairnas} & -0.232 & 91.10 $\pm$ 1.84 & 93.55 & 0.77 & --\\
        ReNAS~\cite{yi2019renas} & 0.634 & 93.90 $\pm$ 0.21 & 94.11 & 0.04 & 423\\
        RegressionNAS & 0.430 & 89.51 $\pm$ 4.94 & 93.65 & 0.40 & 423\\
		\midline
		\sexynamenas (Ours) & \textbf{0.751} & \textbf{93.92 $\pm$ 0.18} & \textbf{94.22} & \textbf{0.01} & 423\\
		\bottomline
	\end{tabular}
	}
	}
	\label{tab:nasbench_search}
\end{table*}

In the following, we will discuss the relationship between architecture comparison and ranking problems.
For convenience, we rewrite $\mR(\alpha, w_{\alpha})$ as $\mR_\alpha$. Given a data set $\{\alpha_i, \mR_{\alpha_i}\}_{i=1}^M $,
% pairwise ranking over architectures aims to
we seek to find a ranking function $f: \Omega \times \Omega \rightarrow \mmR$ by optimizing the ranking loss $\ell_{0}(\alpha, \alpha', \mR_\alpha, \mR_{\alpha'}) =  \mathbbm{1}\{(\mR_\alpha \small{-} \mR_{\alpha'}) f(\alpha, \alpha') \small{\leq} 0\}$.
However, directly optimizing $\ell_0$ is non-trivial since the indicator function is non-differential.
To address this, we use a binary cross-entropy loss $\mL(f;\alpha, \alpha', y) = \mmE\left[ \ell(\sigma \circ f(\alpha, \alpha'), y)\right]$ as a surrogate for $\ell_{0}$, where \sexyname uses the sigmoid activation function $\sigma(\cdot)$ for the output layer, and thus can be denoted as $\sigma \circ f$.
We have the following proposition for \sexyname.
\begin{prop}
\label{prop1}
Let $f:~\Omega \times\Omega \rightarrow \mmR$ be some measurable function, 
% $\sigma(\cdot)$ be the sigmoid function, 
architectures $\alpha, \alpha' \in \Omega$.
The surrogate loss $\mL(f;\alpha, \alpha', y) = \mmE\left[ \ell(\sigma \circ f(\alpha, \alpha'), y)\right]$ is consistent with $\mL_0(f;\alpha, \alpha', \mR_\alpha, \mR_\alpha') = \mmE\left[ \mathbbm{1}\{(\mR_\alpha - \mR_\alpha') f(\alpha, \alpha') \leq 0\}\right]$.
\end{prop}
Proposition~\ref{prop1} shows that learning \sexyname is equivalent to optimizing the ranking over architectures. We put the proof in the supplementary. 

\textbf{Advantages of \sexyname over existing evaluation methods.}
\textbf{1) More stable evaluation results:} our \sexyname is able to provide more stable and accurate reward signals by directly comparing architectures instead of estimating the absolute performance, leading to high rank correlation (See results in Sec.~\ref{exp:nasbench}).
\textbf{2) Lower evaluation time cost:} unlike existing methods evaluate architectures by computing accuracy on the validation data, our \sexyname achieves this only by comparing two architecture graphs. Thus, our method is able to greatly reduce the search cost and accelerate the search process (See more discussions in Sec.~\ref{exp:estimation_cost}).
\textbf{3) Lower requirement for training samples:} given $m$ architectures with the corresponding performance, we can construct $\binom{m}{2}=m(m-1)/2$ training pairs to train \sexyname while the regression-based methods only have $m$ training samples.

\subsection{Data Exploration for Training \sexyname}\label{sec:data_exploration}

Note that learning a good \sexyname requires a set of labeled data, \ie, $\{(\alpha_i, \alpha_i', y_i)\}_{i=1}^M$. 
However, we can only obtain a limited number of labeled data due to the limitation of computational cost in practice.
Given a limited training data set, the performance of \sexyname model may deteriorate, leading to training difficulties for architecture search.
Thus, how to efficiently evaluate architectures using \sexyname with limited data preparations is an important problem.

To address this issue, we propose a data exploration method that adopts the sampled architectures during the search as the unlabeled data.
For these unlabeled data, we propose to take the class with maximum probability predicted by \sexyname as its label.
As shown in Algorithm~\ref{alg:data_exploration},
given the latest \sexyname model $\text{\sexyname}(\cdot,\cdot;\varpi)$, the predicted label of the previously unseen architecture pair can be computed by
\begin{equation}\label{eq:pseudo_label}
    \begin{aligned}
        y' = \mathbbm{1}\{\text{\sexyname}(\alpha, \alpha'; \varpi) \geq 0.5 \},
    \end{aligned}
\end{equation}
where $\mathbbm{1}\{\cdot\}$ refers to an indicator function. Here, $y' = 1$ if $\text{\sexyname}(\alpha, \alpha'; \varpi) \geq 0.5$ and  $y' = 0$ otherwise.
However, the predicted label can be noisy since the \sexyname model may produce wrong predictions. To address this, for the $k$-th architecture pair, we evaluate the prediction quality by computing the confidence score:
\begin{equation}
    f_k = \left| \text{\sexyname}(\alpha_k, \alpha'_k;\varpi) \small{-}0.5 \right|.
\end{equation}
In practice, the higher the confidence score is, the more reliable the predicted label will be.
We select the data with predicted labels with top-$K$ confidence scores and combine them with the labeled data to train \sexyname.
To balance these two kinds of data, we set the proportion of the predicted label data to 0.5 (See discussions in Sec.~\ref{exp:ablation_ratio}).
With the increase of unlabelled data during training, our \sexyname is able to improve the generalization ability to unseen architectures.

\begin{table*}[t]
	\caption{Comparisons of the architectures searched/designed by different methods on ImageNet.
	``--'' means unavailable results.
	$^\dagger$ denotes we test the accuracy from the pretrained model in the official repository.
	``Total Time'' includes the time cost of training the supernet/child networks and the search process.
	``\#Queries'' denotes the number of architectures queried from the supernet for the validation accuracy.
	}
	\centering
	{
	\resizebox{0.88\textwidth}{!}{
    \begin{tabular}{c|ccccccc}
    \topline
    \multicolumn{1}{c|}{\multirow{2}[0]{*}{Search Space}} &
    \multicolumn{1}{c}{\multirow{2}[0]{*}{Architecture}} &
    \multicolumn{2}{c}{Test Accuracy (\%)} &
    \multicolumn{1}{c}{\multirow{2}[0]{*}{\#MAdds (M)}} &
    \multicolumn{1}{c}{\multirow{2}[0]{*}{\#Queries (K)}} &
    \multicolumn{1}{c}{Search Time} &
    \multicolumn{1}{c}{Total Time} \\
    \cline{3-4} &  & \multicolumn{1}{c}{Top-1} & \multicolumn{1}{c}{Top-5} &    &   & (GPU days) & (GPU days) \\
    \midline
    % \multirow{4}*{--} & ResNet-18~\cite{resnet} & 69.8 & 89.1 & 1,814 & -- & -- & --\\
    %  & MobileNetV1 ($1\times$)~\cite{howard2017mobilenets} & 70.6 & 89.5 & 569 & --  & -- & --\\
     & MobileNetV2 ($1.4\times$)~\cite{DBLP:conf/cvpr/SandlerHZZC18} & 74.7 & --  & 585 & -- & -- & --\\
     & ShuffleNetV2 ($2\times$)~\cite{ma2018shufflenet} & 73.7 & -- & 524 & -- & -- & --\\
    \midline
    \multirow{2}*{NASNet} & NASNet-A~\cite{zoph2018learning} & 74.0 & 91.6 & 564 & 20  & -- & 1800\\
     & AmoebaNet-A~\cite{real2018regularized} & 74.5 & 92.0 & 555 & 20  & -- & 3150\\
    % One Shot Small~\cite{bender2018understanding} & -- & 74.2 & -- & -- & 3.3 & 16 \\
    \midline
    \multirow{4}*{DARTS} & DARTS~\cite{liu2018darts} & 73.1 & 91.0 & 595 & 19.5  & 4 & 4\\
     & P-DARTS~\cite{chen2019progressive} & 75.6 & 92.6 & 577 & 11.7  & 0.3 & 0.3\\
     & PC-DARTS~\cite{xu2020pcdarts} & 75.8 & 92.7 & 597 & 3.4 & 3.8 & 3.8\\
     & CNAS~\cite{guo2020breaking} & 75.4 & 92.6 & 576 & 100  & 0.3 & 0.3\\
    \midline
    \multirow{11}*{MobileNetV3-like} & MobileNetV3-Large~\cite{howard2019searching} & 75.2 & -- & 219 & -- & -- & --\\
     & FBNet-C~\cite{DBLP:conf/cvpr/WuDZWSWTVJK19} & 74.9 & --  & 375 & 11.5  & 1.8 & 9\\
     & MnasNet-A3~\cite{tan2019mnasnet} & 76.7 & 93.3 & 403 & 8 & -- & --\\
     & ProxylessNAS~\cite{cai2018proxylessnas} & 75.1 & 92.3  & 465 & --  & -- & 8.3\\
    %  & SPOS~\cite{DBLP:journals/corr/abs-1904-00420} & 74.4 & 91.8 & 323 & 1  & \textless1 & 12\\
    % FairNAS~\cite{chu2019fairnas} & 77.5 & 93.7 & 5.9 & 392 & 2 & 10 \\
    %  & OFA~\cite{Cai2020Once} & 76.0 & --  & 230 & 1.7 & 51.7\\
     & OFA~\cite{Cai2020Once} & 76.0 & --  & 230 & 16 & 1.7 & 51.7\\
    % DNA~\cite{DNA}& 78.4 & 94.0 & 6.4 & 611 & - \\
    %  & FBNetV2~\cite{FBNETV2} & 77.2 & --  & 325 & 11.5 & 5 & 25\\
    & FBNetV2~\cite{FBNETV2}$^\dagger$ & 76.3 & 92.9  & 321 & 11.5 & 5 & 25\\
     & AtomNAS~\cite{Mei2020AtomNAS} & 75.9 & 92.0  & 367 & 78 & -- & --\\
    % \midline
     & Random Search & 76.0 & 92.6  & 314 & 1  & -- & 50\\
     & Best Sampled Architectures & 76.7 & 93.1 & 382 & 1  & -- & 50\\
     & \sexynamenas (Ours) & \textbf{77.3} & \textbf{93.4} & 482 & 1 & \textbf{0.1} & 50.1\\
    \bottomline
    \end{tabular}
    }
    }
	\label{tab:imagenet}
\end{table*}

\section{Experiments}

We apply our \sexynamenas to three different search spaces, namely NAS-Bench-101~\cite{ying2019bench},  MobileNetV3~\cite{howard2019searching} and DARTS~\cite{liu2018darts}\footnote{Due to page limit, we put the experimental results in DARTS search space into the supplementary.} search spaces.
We put more details about the search space and the implementation in supplementary.
All implementations are based on PyTorch\footnote{The source code is available at \href{https://github.com/chenyaofo/CTNAS}{https://github.com/chenyaofo/CTNAS}.}.

\subsection{Experiments on NAS-Bench-101}\label{exp:nasbench}

\textbf{Implementation Details}.
For a fair comparison, we set the number of architecture-accuracy pairs queried from the NAS-Bench-101 dataset to 423 for all the methods.
To measure the rank correlation, we use another 100 architectures in the dataset to compute Kendall's Tau (KTau)~\cite{kendalltau}.
Note that a larger KTau means evaluating architectures more accurately.
Following the settings in ~\cite{ying2019bench}, we obtain the test accuracy by averaging the accuracy of 3 different runs.
Following~\cite{pham2018efficient}, we train the \sexynamenas model for 10k iterations with a batch size of 1 in the training.
% More implementation details are put into the supplementary.

\begin{figure}[t]
\centering
\includegraphics[width=0.9\linewidth]{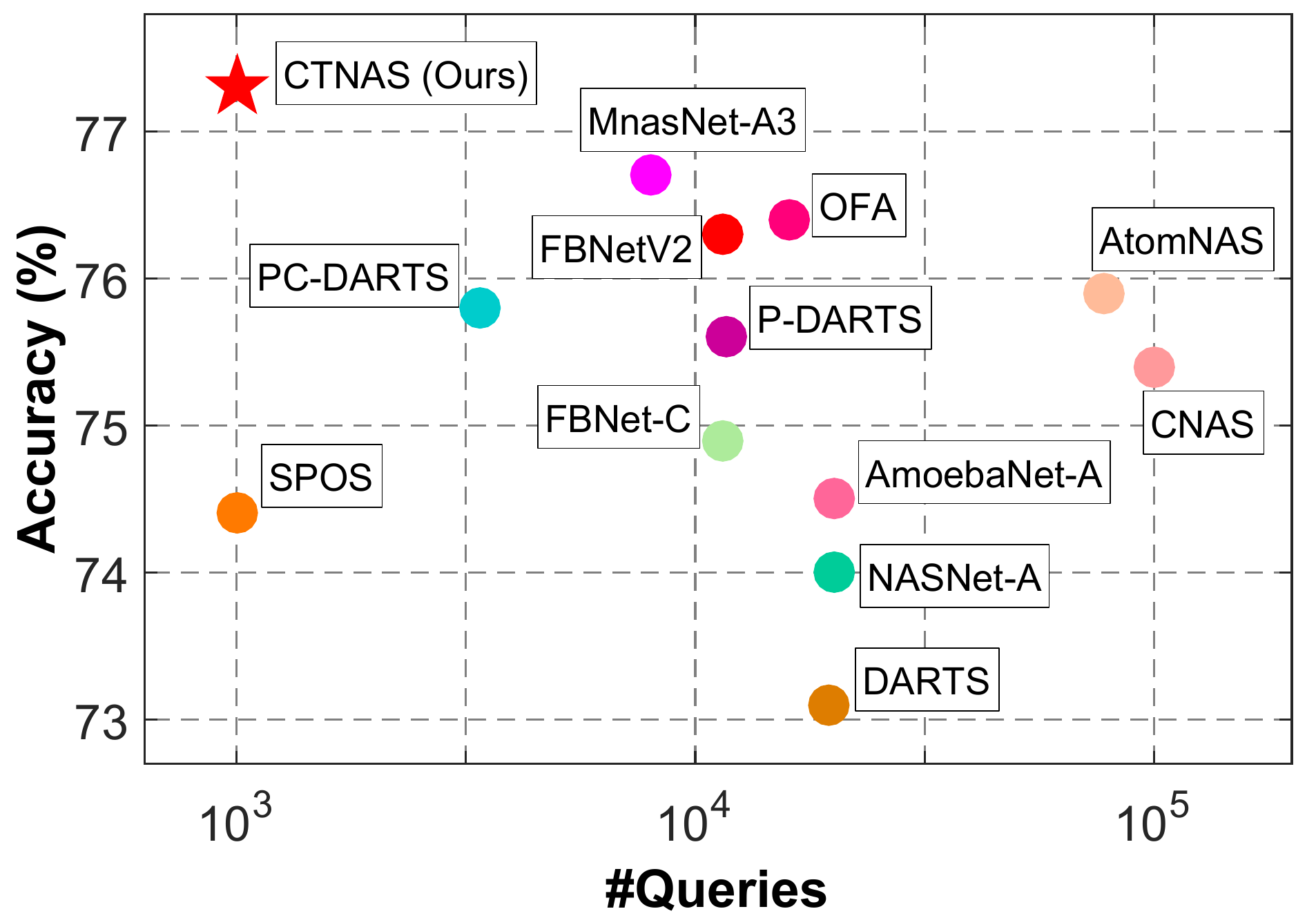}
\caption{
The accuracy \vs~the number of queries among different methods on ImageNet.
}
\label{fig:comparisons_sota}
\end{figure}

\noindent\textbf{Comparisons with State-of-the-art Methods}.
We compare our proposed \sexynamenas with state-of-the-art methods in NAS-Bench-101 search space.
From Table~\ref{tab:nasbench_search}, the proposed \sexynamenas achieves higher KTau value (0.751) than other NAS methods.
The results show \sexynamenas evaluates architectures accurately, which is beneficial to the search process.
Besides, \sexynamenas achieves the highest average testing accuracy (93.92\%), and the best architecture searched by \sexynamenas is the top 0.01\% in the search space with the testing accuracy of 94.22\%.
We also compare \sexynamenas with its variant (\ie, RegressionNAS) that trains a regression-based predictor with L2 loss to predict the absolute performance.
\sexynamenas has a much higher average testing accuracy and a lower variance of the accuracy than RegressionNAS, which demonstrates the stability of our method.

\subsection{Experiments on ImageNet}\label{exp:imagenet}

\textbf{Implementation Details}.
We apply our \sexynamenas to a MobileNetV3-like search space~\cite{howard2019searching}.
We first train a supernet with the progressive shrinking strategy~\cite{Cai2020Once} (cost 50 GPU days).
Then, we sample 1000 architectures in the search space as the training data for \sexyname.
Following~\cite{Cai2020Once}, we compute the validation accuracy on 10k images sampled from the training set of ImageNet.
For a fair comparison, we following the mobile setting~\cite{liu2018darts} and restrict the number of multiply-adds (MAdds) to be less than 600M. 
% We put more implementation details in the supplementary.

\noindent\textbf{Comparisons with State-of-the-art Methods}.
We compare the performance of \sexynamenas with other NAS methods on ImageNet in Table~\ref{tab:imagenet}.
\sexynamenas achieves $77.3\%$ top-1 accuracy and 93.4\% top-5 accuracy, which consistently outperforms existing human-designed architectures and state-of-the-art NAS models searched in different search spaces.
Besides, we compare the searched architecture with the best one of 1000 sampled architectures.
The searched architecture achieves higher accuracy (77.3\% \vs~76.7\%), which demonstrates the effectiveness of \sexynamenas.
We report the number of queries from supernet for validation accuracy of different methods in Figure~\ref{fig:comparisons_sota}.
Note that querying from supernet takes up most of the search time cost.
Thus, the number of queries becomes an important metric to measure search cost~\cite{ying2019bench,luo2020seminas}.
\sexynamenas has fewer queries (1k) than other methods but achieves the highest searched performance.
The results demonstrate that \sexyname is more efficient and greatly accelerates the search process (\ie, only 0.1 GPU days).

\section{Further Experiments}

\subsection{Comparisons of Architecture Evaluation Cost}\label{exp:estimation_cost}
In this experiment, we compare our \sexyname with other NAS methods in terms of the time cost to rank 100 neural architectures.
From Table~\ref{tab:estimation_time_cost}, our \sexyname has much lower time cost (4.1 ms) than existing NAS methods, such as ReNAS~\cite{yi2019renas} (85.6 ms) and ENAS~\cite{pham2018efficient} (2.7 s).
The reasons have two aspects: 
1) The inputs of \sexyname have lower dimensions. \sexyname only takes architecture graphs as inputs while weight sharing methods (\eg, ENAS) need to compute the accuracy on the validation data. Besides, ReNAS uses manually-deigned features of architectures as inputs which have higher dimensions than ours.
2) Our \sexyname only consists of 3 layers while the models in considered methods often have tens of layers.

\begin{table}[H]
	\caption{Time cost of evaluating 100 architectures.}
	\centering
    {
    \resizebox{\linewidth}{!}{
	\begin{tabular}{c|cccc}
    	\topline
        Method & \sexynamenas & ReNAS & ENAS & Training from Scratch\\
    	\midline
    	Time Cost & \textbf{4.1 ms} & 85.6 ms & 2.7 s & 5,430 h\\
    	\bottomline
	\end{tabular}
	}
	}
	\label{tab:estimation_time_cost}
\end{table}

\subsection{Effect of Baseline Updating Scheme}\label{exp:baseline_updating}

To verify the effectiveness of the proposed baseline updating scheme via curriculum learning, we compare our \sexynamenas with two variants, namely \textit{Fixed Baseline} and \textit{Randomly Updating}.
Fixed Baseline variant finds promising architectures with a fixed baseline architecture in the whole searching process.
Random Updating variant updates the baseline with a randomly sampled architecture.
From Table~\ref{tab:diff_baseline_updating}, our \sexynamenas outperforms these two variants by a large margin.
The reason is that the fixed or randomly sampled baseline may be too weak. In this case, it is hard for the controller to find promising architectures.
\cyf{Randomly sampled variant samples sufficient good baselines in some cases, thus achieves higher performance than the fixed one.}

\begin{table}[H]
	\caption{Comparisons of different baseline updating schemes.}
	\centering
    {
    \resizebox{1.0\linewidth}{!}{
	\begin{tabular}{c|ccc}
		\topline
		Method & \sexynamenas & Fixed Baseline & Random Updating  \\
		\midline
		Acc. (\%) & \textbf{93.92$\pm$0.18} & 93.39$\pm$0.17 & 93.53$\pm$0.39 \\
		\bottomline
	\end{tabular}
	}
	}
	\label{tab:diff_baseline_updating}
\end{table}

\subsection{Effect of Data Exploration Method}\label{exp:ablation_ratio}

To investigate the effect of the proposed data exploration method, we conduct more experiments in NAS-Bench-101 search space.
Let $\mM$ and $\mU$ be the set of labeled data and data with labels predicted by \sexyname in a batch, respectively.
The proportion of data with predicted labels is $r=|\mU|/(|\mU|+|\mM|)$.
As shown in Table \ref{tab:ratio},
compared with that without data exploration ($r=0$), \sexynamenas achieves better performance when the proportion $r$ is set to 0.5.
Besides, either a low or high proportion would hamper the training of \sexynamenas, leading to a poor searched performance.
Thus, we set the proportion $r$ to 0.5 in the experiments.

\begin{table}[H]
	\caption{Comparisons of different proportions ($r$) of data with predicted labels.}
	\centering
% 	\vspace{-2.25pt}
    {
    \resizebox{1.0\linewidth}{!}{
	\begin{tabular}{c|cccc}
		\topline
		$r$ & 0.0 & 0.3 & 0.5 & 0.8 \\
		\midline
		Acc. (\%) & 93.60$\pm$0.22& 93.62$\pm$0.30& \textbf{93.92$\pm$0.18}& 93.60$\pm$0.40 \\
		\bottomline
	\end{tabular}
	}
	}
	\label{tab:ratio}
\end{table}

\subsection{Effect of the Number of Training Samples}\label{exp:ablation_sample}
\begin{figure}[H]
\centering
\includegraphics[width=0.85\linewidth]{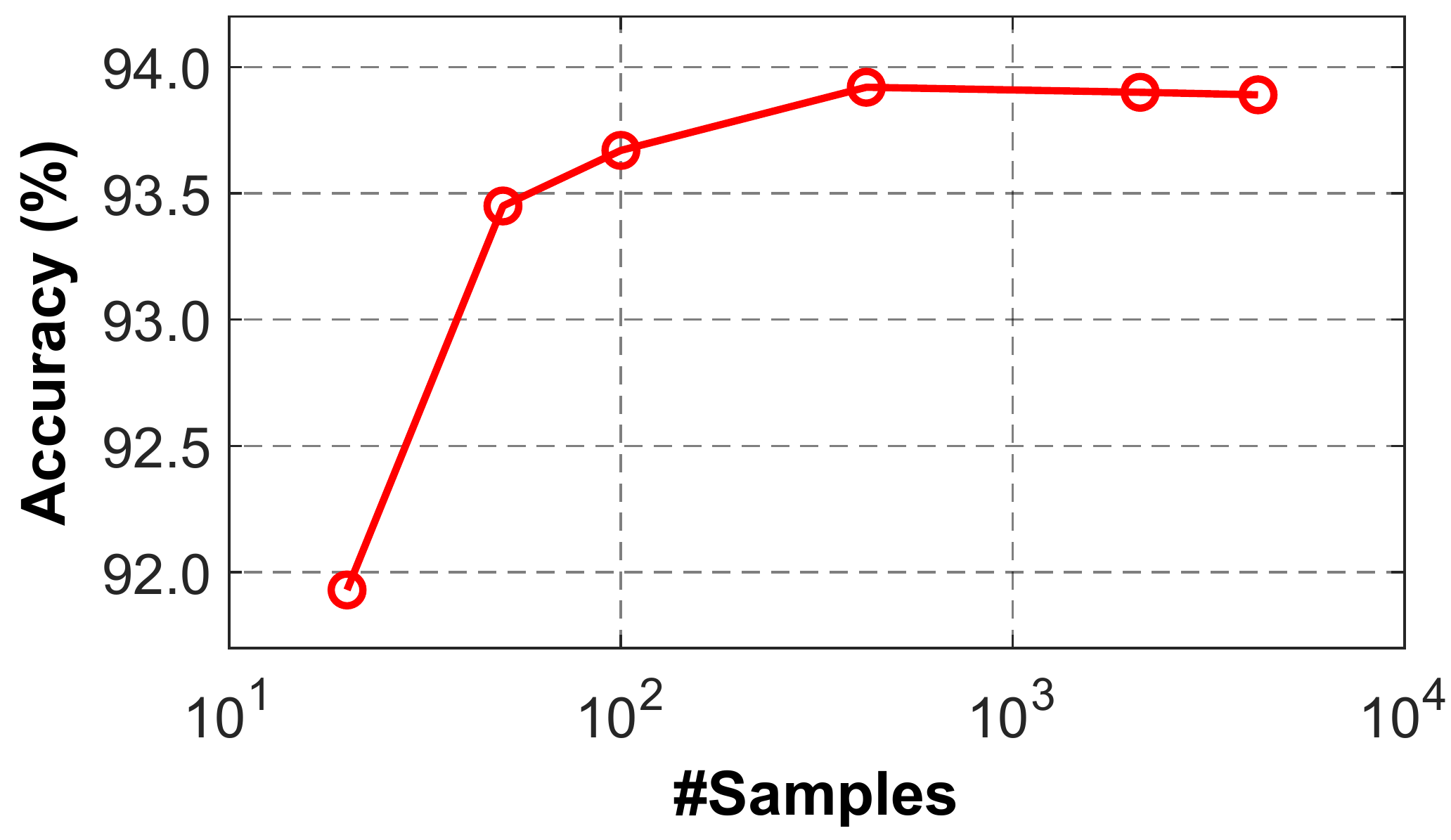}
\caption{
Comparisons of different number of training samples.
}
\label{fig:num_training_samples}
\end{figure}

To investigate the effect of the number of training architectures/samples, we perform more experiments with different numbers of training samples (\ie, \{20, 50, 100, 423, 2115, 4230\}).
% We report the average test accuracy of 10 different runs.
From Figure~\ref{fig:num_training_samples},
when the number of samples increases from 20 to 423, our \sexynamenas achieve better performance.
However, our \sexynamenas yield similar accuracy when the number of samples is larger than 423.
The results show that a small number of training samples are sufficient for our \sexyname.
The reasons are two folds:
1) The proposed data exploration method provides more training samples.
2) Our \sexyname has a low requirement for training samples since we can construct $m(m-1)/2$ pairs from $m$ labeled samples.

\section{Conclusion}

In this paper, we have proposed a novel Contrastive Neural Architecture Search (\sexynamenas) method that performs search based on the comparison results of the sampled architecture and a baseline one.
To constantly find better architectures, we propose a baseline updating scheme via curriculum learning to improve the baseline gradually.
To provide the comparison results, we devise a Neural Architecture Comparator (\sexyname) that takes two architectures as inputs and outputs the probability that one is better than the other.
Moreover, we propose a data exploration method for \sexyname to reduce the requirement of training data and improve the generalization ability of \sexyname to evaluate unseen architectures.
The experimental results in three search spaces demonstrate that our \sexynamenas outperforms the state-of-the-art methods.

\vspace{-12pt}

\cyf{
{\flushleft \bf Acknowledgements}.
This work was partially supported by Key-Area Research and Development Program of Guangdong Province (2019B010155002),
CCF-Baidu Open Fund (CCF-BAIDU OF2020022), National Natural Science Foundation of China (NSFC) 61836003 (key project), Program for Guangdong Introducing Innovative and Enterpreneurial Teams 2017ZT07X183, Fundamental Research Funds for the Central Universities D2191240.
Yong Guo is supported by National Natural Science Foundation of China (Grant No. 62072186), Guangdong Basic and Applied Basic Research Foundation (Grant No. 2019B1515130001).
}

% \clearpage
{\small
\bibliographystyle{ieee_fullname}
\bibliography{egbib}
}

% \eat
{
\clearpage
\onecolumn
\appendix

\begin{center}
	{
		\Large{\textbf{Supplementary for \\ ``\mytitle''}}
	}
\end{center}
\vspace{15 pt}

In the supplementary, we provide a detailed proof of Proposition 1, more experimental results and more details on the \sexynamenas method.
We organize our supplementary as follows.

\begin{itemize}
	\item In Section~\ref{supp:sec:theoretical_proofs}, we provide the detailed proof for Proposition~1.
	\item In Section~\ref{supp:sec:training_data}, we describe how to construct the training data for \sexynamenas.
	\item In Section~\ref{supp:sec:training_details}, we show more training details of \sexynamenas.
	\item In Section~\ref{supp:sec:evaluation_architectures}, we detail the evaluation method for the searched architectures.
	\item In Section~\ref{supp:sec:search_space}, we provide more details on the three considered search spaces. 
	\item In Section~\ref{supp:sec:evaluation_estimators}, we give more details on the evaluation metric Kendall's Tau.
	\item In Section~\ref{supp:sec:exp_darts}, we give more experimental results on DARTS search space. 
    \item In Section~\ref{supp:sec:visualization}, we show the visualization results of the searched architectures.
\end{itemize}

\section{Proof of Proposition 1}\label{supp:sec:theoretical_proofs}
In this section, we provide the detailed proof of Proposition~\ref{prop1}. We first introduce a useful theorem.

Let $\mX$ be a set of samples to be ranked. Given $x,x' {\in} \mX$ with the labels $\mR_x$ and $\mR_{x'}$, a score function $f: \mX \times \mX \rightarrow \mR$, the pairwise ranking loss is defined as $\ell_0(x,x',\mR_x, \mR_{x'}) = \mathbbm{1}{\{(\mR_x\small{-}\mR_{x'}) f(x, x') \small{\leq} 0\}}$. The expected true risk is defined as $R_0(f)=\mmE\left[\ell_0 \right]$. Since the indicator function is non-differential, directly optimize $\ell_0$ is non-trivial. In practice, the pairwise ranking problem aims to learn a ranking function by optimizing a surrogate loss function.
Let $\Phi(\cdot)$ be some measurable function. A surrogate loss can be defined as $\ell_\Phi(x,x',\mR_x, \mR_{x'}) = \Phi(- \text{sgn}(\mR_x - \mR_{x'}) \cdot f(x, x'))$.
The expected surrogate risk is defined as $R_\Phi(f) = \mmE\left[\ell_\Phi \right]$. The following theorem~\cite{liu2009learning} 
analyzes the statistical consistency of the surrogate loss $\ell_\Phi$ with respect to the true loss $\ell_0$. 

\begin{thm}
	\label{thm1}
Let $R_0^*=\inf_f R_0(f)$ and $R_\Phi^*=\inf_f R_\Phi(f)$. Then for all functions $f$, as long as $\Phi$ is convex, we have
\begin{equation}\label{ineqn}
    R_0(f) - R_0^* \leq \Psi^{-1}(R_\Phi(f) - R_\Phi^*),
\end{equation}
where $$\Psi(x) = H^-(\frac{1 + x}{2}) - H^-(\frac{1 - x}{2}),$$$$
H^-(\rho) = \inf_{\alpha:\alpha(2\rho - 1) \leq 0}(\rho \Phi(-\alpha) + (1 - \rho)\Phi(\alpha)).$$
\end{thm}

The above theorem states that if an algorithm can effectively minimize the right-hand side of the inequality (\ref{ineqn}) to approach zero, then the left-hand side will also approaches zero. In other words, if $\Phi(\cdot)$ is a convex function, the surrogate loss $\ell_\Phi$ is statistically consistent with the true loss $\ell_0$. Based on Theorem~\ref{thm1}, we then prove Proposition~\ref{prop1}.

\flushleft\textbf{Proposition 1}
\textit{
Let $f:~\Omega \times\Omega \rightarrow \mR$ be some measurable function, $\sigma(\cdot)$ be the sigmoid function, and $\alpha, \alpha' \in \Omega$. The surrogate loss $\mL(f;\alpha, \alpha', y) = \mmE\left[ \ell(\sigma \circ f(\alpha, \alpha'), y)\right]$ is consistent with $\mL_0(f;\alpha, \alpha', \mR_\alpha, \mR_{\alpha'}) = \mmE\left[ \mathbbm{1}\{(\mR_\alpha - \mR_{\alpha'}) f(\alpha, \alpha') \leq 0\}\right]$.
}
  
   \textbf{Proof} Given $y = \mathbbm{1}\{\mR_\alpha - \mR_{\alpha'} \ge 0\}$ and $\sigma(x) = \frac{1}{1 + e^{-x}}$, we have the following equation,
  	\begin{equation}
	\begin{aligned}
	\mL(f;\alpha, \alpha', y) 
	= &\mmE\left[ - y\log \sigma(f(\alpha, \alpha')) - (1 - y) \log (1 - \sigma(f(\alpha, \alpha'))\right] \\
	= &\mmE\left[ -y\log \sigma(f(\alpha, \alpha')) + (1 - y) (f(\alpha, \alpha') - \log \sigma(f(\alpha, \alpha')))\right] \\
	= &\mmE\left[(1 - y)f(\alpha, \alpha') + \log (1 + \exp(-f(\alpha, \alpha')))\right]  \\
	= &\mmE\left[ \log(1 + \exp(-\text{sgn}(\mR_\alpha - \mathcal{R}_{\alpha'}) \cdot f(\alpha, \alpha')))\right] \\
	= &\mmE\left[\Phi (-\text{sgn}(\mR_\alpha - \mR_{\alpha'}) \cdot f(\alpha, \alpha'))\right],
	\end{aligned}
	\end{equation}
where $\Phi(x) = \log(1+\exp(x))$. Since $\Phi(x)$ is a convex function, according to Theorem \ref{thm1}, the surrogate loss $L(f;\alpha, \alpha', y)$ is consistent with $L_0(f;\alpha, \alpha', \mR_\alpha, \mR_{\alpha'})$.

\section{More Details on Data Construction}\label{supp:sec:training_data}

As mentioned in Section 4, the training of \sexyname relies on a set of training data. Here, we provide more details on the training data of NAS-Bench-101 search space~\cite{ying2019bench}, MobileNetV3-like search space~\cite{howard2019searching}
and DARTS search space~\cite{liu2018darts}.

\textbf{Data Construction in NAS-Bench-101 Search Space.}
To train the proposed \sexynamenas model, we use the architectures and the corresponding performance as training data in the NAS-Bench-101 search space.
Google has released $423$k architectures with corresponding performance on CIFAR-10 in the NAS-Bench-101 search space.
The performance includes the training accuracy, validation accuracy, test accuracy, training time and the number of trainable parameters.
We randomly sample $423$ architectures ($0.1\%$ of the whole architectures) as the training set and 100 architectures as the validation set.
For training the proposed \sexynamenas, we construct the training architecture pair $(\alpha, \alpha')$ by randomly sampling from the training architecture set.
The label of the architecture pair $(\alpha, \alpha')$ is computed from the validation accuracy of architecture $\alpha$ and $\alpha'$.

\textbf{Data Construction in MobileNetV3-like Search Space.} To obtain the training data pairs for \sexynamenas in the MobileNetV3-like search space, we train a supernet with a progressive shrinking strategy following the setting in~\cite{Cai2020Once}.
We then randomly sample 1000 architectures in the search space and  apply the weights from the supernet ($1.0 \times$) to these architectures.
Then we evaluate the architectures on 10000 validation images sampled from the training set of ImageNet to obtain the validation accuracy.
The evaluation takes less than 0.1 GPU days.
Finally, we compute the label of the training pairs for \sexyname with the validation accuracy of the architectures.

\textbf{Data Construction in DARTS Search Space.}
Training \sexynamenas requires some architectures with the corresponding performance.
To this end, we first train a supernet for 120 epochs with a uniform sampled strategy~\cite{DBLP:journals/corr/abs-1904-00420}.
Following DARTS~\cite{liu2018darts}, we train the supernet on $50\%$ of the official training set of CIFAR-10 and evaluate them for the validation accuracy on the remain $50\%$ using standard data augmentation techniques~\cite{resnet}.
The batch size and initial learning rate are set to 96 and 0.01, respectively.
Then, we randomly sample $1000$ architectures in the search space.
The validation accuracy of each sampled architecture is computed by inheriting weights from the supernet.
The label of the training pairs for \sexyname is computed from the average validation accuracy of the last 10 epochs in the supernet training.

\section{More Training Details of \sexynamenas}\label{supp:sec:training_details}
In this section, we show more training details of \sexynamenas while searching in the NAS-Bench-101~\cite{ying2019bench} and MobileNetV3-like search space~\cite{howard2019searching}.
We follow the settings in~\cite{kipf2016semi} to build our NAC model. Specifically, we first use two GCN layers to extract features of two input architecture. Then, we concatenate the features and send them to a fully-connected (FC) layer to produce the probability of the first architecture being better than the other one. 
Since the number of nodes of the architecture graph may be less than $7$ in NAS-Bench-101 search space, we pad the size of the adjacency matrix to $7$ with zero paddings.

In the training, we train the \sexynamenas model for 10k iterations.
We use Adam with a learning rate of $2\times10^{-4}$ and a weight decay of $5\times10^{-4}$ as the optimizer.
The training batch size of \sexyname is set to 256.
To explore more training data, we randomly sample 512 architecture pairs every iteration and add the top 256 architecture pairs with the predicted labels according to comparison probability into the training data batch (See Algorithm~3).
We update the baseline architecture every 1000 iteration.
We add the controller’s sample entropy to the reward, which is weighted by $5\times10^{-4}$.

\section{More Details on Architecture Evaluation}\label{supp:sec:evaluation_architectures}

In this section, we elucidate the details of evaluation method for the searched architectures in NAS-Bench-101~\cite{ying2019bench}, MobileNetV3-like~\cite{howard2019searching} and DARTS~\cite{liu2018darts} search space.

\textbf{More Evaluation Details in NAS-Bench-101 search space.}
NAS-Bench-101 has released the performance of all the architectures in its search space, including the training accuracy, the validation accuracy and the test accuracy. Following the settings in NAS-Bench-101, we report the average test accuracy of the searched architecture over 3 different runs.

\textbf{More Evaluation Details in MobileNetV3-like search space.}
We evaluate the architectures searched in MobileNetV3-like search space on ImageNet.
Following the setting in~\cite{liu2018darts}, the number of multiply-adds (MAdds) in the searched architectures is restricted to be less than $600$M.
Following the setting in~\cite{Cai2020Once}, we directly apply the weights form the supernet (1.0$\times$) to the searched architectures, and then evaluate them on the validation set of ImageNet~\cite{deng2009imagenet} without finetune.

\textbf{More Evaluation Details in DARTS search space.}
To evaluate the searched architectures, we train them from scratch on CIFAR-10.
We build the final convolution network with $18$ learned cells, including 16 normal cells and $2$ reduction cells.
The two reduction cells are put at the $1/3$ and $2/3$ depth of the network, respectively.
The initial number of the channels is set to $44$.
Following the setting in~\cite{liu2018darts},
we train the model for $600$ epochs using the batch size of $96$.
The training images are padded $4$ pixels on each side.
Then the padded images or their horizontal flips are randomly cropped to the size of $32 \times 32$.
We use cutout~\cite{devries2017improved} with a length of $16$ in the data augmentation.
We use an SGD optimizer with a weight decay of $3 \times 10^{-4}$ and a momentum of $0.9$.
The learning rate starts from $0.025$ and follows the cosine annealing strategy with a minimum of $0.001$.
Additional enhancements include path dropout~\cite{DBLP:conf/iclr/LarssonMS17} of probability of $0.2$ and auxiliary towers~\cite{szegedy2015going} with a weight of $0.4$.

\section{More Details on Search Space}\label{supp:sec:search_space}
In this paper, we consider three search spaces, namely the NAS-Bench-101 search space~\cite{ying2019bench}, MobileNetV3-like search space~\cite{howard2019searching} and DARTS search space~\cite{liu2018darts}.
We show the details of these search spaces below.

\textbf{NAS-Bench-101 Search Space.}
NAS-Bench-101 defines a search space that is restricted to small topology structures, usually called cells.
Each cell contains $7$ nodes at most, including IN and OUT nodes, which represent the input and output tensors of the cell.
The remaining nodes can be selected from $3 \times 3$ convolution, $1\times 1$ convolution and $3 \times 3$ max pooling.
The number of connections in a cell is limited to no more than $9$.
The whole convolution network is stacked with $3$ blocks followed by a global average pooling and a fully-connected layer.
Each block is stacked with $3$ cells followed by a downsampling layer, where the image height and width are halved via max-pooling and the channel count is doubled.
The initial layer of the model is a stem consisting of a $3\times3$ convolution with $128$ output channels.

\textbf{MobileNetV3-like Search Space.}
We consider the search space proposed by MobileNetV3.
The model is divided into 5 units with gradually reduced feature map size and increased channel number.
Each unit consists of 4 layers at most where only the first layer has stride 2 if the feature map size decrease.
All the other layers in the units have stride 1.
In our experiments, we search for the number of layers in each unit (chosen from $\{2,3,4\}$), the kernel size in each layer (chosen from $\{3,5,7\}$) and the width expansion ratio in each layer (chosen from $\{3,4,6\}$).

\textbf{DARTS Search Space.}
We consider a cell-based search space proposed by DARTS~\cite{liu2018darts},
which includes two cell types, namely the normal cell and reduction cell.
The normal cell keeps the spatial resolution of feature maps.
The reduction cell reduces the height and width of feature maps by half and doubles the number of channels.
Each cell contains $7$ nodes, including 2 input nodes, 4 intermediate nodes and 1 output node. 
The output node is the concatenation of the 4 intermediate nodes.
There are $8$ candidate operations between two nodes, including $3 \times 3$ depthwise separable convolution, $5 \times 5$ depthwise separable convolution, $3 \times 3$ max pooling, $3 \times 3$ average pooling, $3 \times 3$ dilated convolution, $5 \times 5$ dilated convolution, identity, and none.

\section{More Details on Kendall's Tau}\label{supp:sec:evaluation_estimators}
To evaluate the performance estimators, we use Kendall's Tau (KTau)~\cite{kendalltau} to compute the ranking correlation between the predicted performance and the performance obtained by training from scratch over a set of architectures.
Given $n$ architectures $\{\alpha_i\}_{i=1}^n$, there are ${{n}\choose{2}}=\frac{n(n-1)}{2}$ architectures pairs.
The KTau $\tau$ can be computed by
\begin{equation}
\label{eq:ktau}
    \begin{aligned}
    \tau = {{n}\choose{2}}^{-1}
    \sum_{i<j}\big[
    \text{sgn}(f(\alpha_i) - f(\alpha_j))
    \cdot \\
    \text{sgn}(\mathcal{R}(\alpha_i,w_{\alpha_i}) - \mathcal{R}(\alpha_j,w_{\alpha_j}))
    \big],
    \end{aligned}
\end{equation}
where $\text{sgn}(\cdot)$ is a sign function, $f(\cdot)$ is a performance predictor, $\mathcal{R}(\cdot, w)$ is the performance obtained by training from scratch.
% The KTau $\tau$ is in the range of $[-1,1]$.
The value of $\tau$ represents the correlation between the ranking of the predicted performance and the performance obtained by training from scratch.
A higher KTau means the predicted performance ranking is more accurate.

We compare the KTau of different NAS methods in the NAS-Bench-101 search space. The results are shown in the Table~\ref{tab:comparisons_ktau}.
Our proposed \sexynamenas achieves higher KTau ($0.751$) than the considered methods, including ReNAS~\cite{yi2019renas}, SPOS~\cite{DBLP:journals/corr/abs-1904-00420} and FairNAS~\cite{chu2019fairnas}.
These results mean that the performance ranking predicted by our method is very close to that of training from scratch.
In the search process, the higher KTau means a more accurate reward, often resulting in better search performance.

\begin{table}[h]
\renewcommand\thetable{A}
	\centering
	\caption{Comparisons of the KTau in the NAS-Bench-101 search space for different NAS methods.}
	\label{tab:comparisons_ktau}
    {
    \resizebox{0.55\linewidth}{!}{
	\begin{tabular}{ccccc}
		\hline
        Method &  ReNAS~\cite{yi2019renas} & SPOS~\cite{DBLP:journals/corr/abs-1904-00420} & FairNAS~\cite{chu2019fairnas}  & \sexynamenas (Ours)  \\
		\hline
		KTau & 0.634 & 0.195 & -0.232 & \textbf{0.751} \\
		\hline
	\end{tabular}
	}
	}
\end{table}

\section{More Results on DARTS Search Space}\label{supp:sec:exp_darts}

\textbf{Implementation Details}.
We train a supernet for 120 epochs with a uniform sampled strategy~\cite{DBLP:journals/corr/abs-1904-00420}.
We randomly sample $1000$ architectures in the DARTS search space~\cite{liu2018darts}, and obtain their validation accuracy by inheriting weights from the supernet to construct the data for \sexyname training.
We train the controller for 10k iteration with a batch size of 1 following the settings in~\cite{pham2018efficient}.
We add the controller’s sample entropy to the reward, which is weighted by $5\times10^{-4}$.

\textbf{Comparisons with State-of-the-art Methods}.
We compare the searched performance of our proposed \sexynamenas with the state-of-the-art methods in DARTS~\cite{liu2018darts} search space.
From Table~\ref{tab:cifar-10},
the architecture searched by \sexynamenas outperforms both manually designed and automatically searched architectures.
We also compare the searched architecture with the best one in 1000 sampled architectures. The searched architecture achieves higher average accuracy (97.41\% \vs~97.33\%), which demonstrates the effectiveness of the proposed \sexynamenas.
Moreover, our \sexynamenas takes 0.2 GPU days to train the supernet and 0.1 GPU day for the search phrase, which is much faster than the considered NAS methods.
The reason is that our \sexyname has a much lower time cost while evaluating architectures (See results in Sec.~6.1 in the paper).

\begin{table}[H]
\renewcommand\thetable{B}
	\caption{Comparisons with
	state-of-the-art models on CIFAR-10.
	``cutout'' indicates evaluating the architecture using cutout~\cite{devries2017improved} data argumentation.
	``--'' means unavailable results.
	}
	\centering
	\resizebox{0.76\textwidth}{!}{
    	\label{tab:cifar-10}
    	\begin{tabular}{cccc}
    		\topline
    		Architecture & Test Accuracy (\%)& \#Params. (M) & Search Cost (GPU days) \\
    		\midline
    		DenseNet-BC~\cite{huang2017densely}&96.54&25.6 & --\\
    		PyramidNet-BC~\cite{han2017deep}&96.69&26.0 & --\\
    		\hline
    		Random search baseline &96.71 $\pm$ 0.15 &3.2 & --\\
    		NASNet-A + cutout~\cite{zoph2018learning}&97.35&3.3 & 1,800\\
    		NASNet-B~\cite{zoph2018learning}&96.27&2.6 & 1,800\\
    		NASNet-C~\cite{zoph2018learning}&96.41&3.1 & 1,800\\
    		AmoebaNet-A + cutout~\cite{real2018regularized}&96.66 $\pm$ 0.06&3.2 &3,150\\
    		AmoebaNet-B + cutout~\cite{real2018regularized}&96.63 $\pm$ 0.04&2.8 &3,150\\
    % 		Hierarchical Evo~\cite{liu2017hierarchical}&96.25 $\pm$ 0.12 &15.7 &300\\
    		SNAS~\cite{xie2018snas}&97.02&2.9 &1.5\\
    		ENAS + cutout~\cite{pham2018efficient}&97.11&4.6 &0.5\\
    		NAONet~\cite{luo2018neural}&97.02&28.6 &200\\
    % 			NAONet-WS~\cite{luo2018neural}&96.47&2.5 &--&0.3\\
    % 		NAT-DARTS~\cite{guo2019nat}&97.28&2.7&--\\
    		GHN~\cite{zhang2018graph} & 97.16 $\pm$ 0.07 & 5.7 & 0.8\\
    		PNAS + cutout~\cite{DBLP:conf/eccv/LiuZNSHLFYHM18} & 97.17 $\pm$ 0.07 &3.2& --\\
    		DARTS + cutout~\cite{liu2018darts} &97.24 $\pm$ 0.09 &3.4 &4\\
    		\hline
    		Best in Sampled Architectures + cutout & 97.33 $\pm$ 0.07 & 3.7 & --\\
    		\sexynamenas + cutout (Ours) & \textbf{97.41 $\pm$ 0.04} & 3.6 &\textbf{0.3}\\
    		\bottomline
    	\end{tabular}
        }
\end{table}

\section{Visualization Results of Searched Architectures}\label{supp:sec:visualization}
We show the visualization results of searched architectures of three consider search space in Figure~\ref{fig:nasbench_searched_arch},~\ref{fig:nacnas_mb_arch} and~\ref{fig:cnn_arch}, respectively.
In Figure~\ref{fig:nasbench_searched_arch}, we show the best architecture searched by \sexynamenas in NAS-Bench-101 search space, which is the third best architecture in the whole search space.
In Figure~\ref{fig:nacnas_mb_arch}, we show the resulting architecture searched in MobileNetV3-like search space, which achieves 77.3\% top-1 accuracy and 93.4\% top-5 accuracy on ImageNet. These architectures searched by \sexynamenas outperform existing human-designed architectures and automatically searched architectures.
The visualization results of the normal cell and reduction cell searched by \sexynamenas in DARTS search space are shown in Figure~\ref{fig:cnn_arch}, which achieves the average accuracy of 97.41\% on CIFAR-10.

\begin{figure}[H]
\renewcommand\thefigure{A}
\centering
\includegraphics[width=0.4\columnwidth]{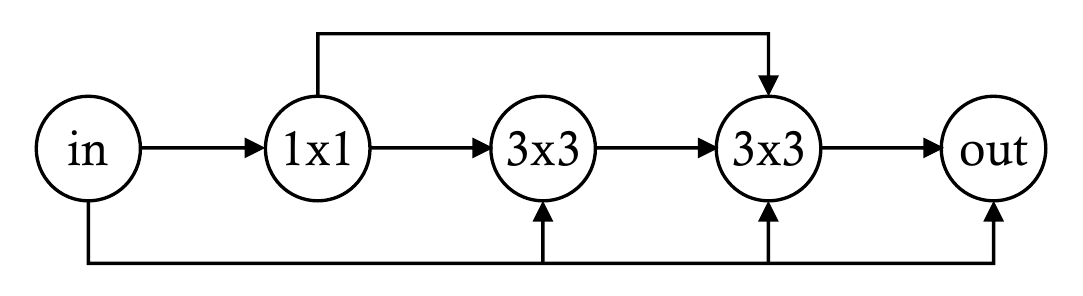}
\caption{
    The best architecture searched by \sexynamenas in NAS-Bench-101 search space.
}
\label{fig:nasbench_searched_arch}
\end{figure}

\begin{figure}[H]
\renewcommand\thefigure{B}
\centering
\includegraphics[width=0.8\columnwidth]{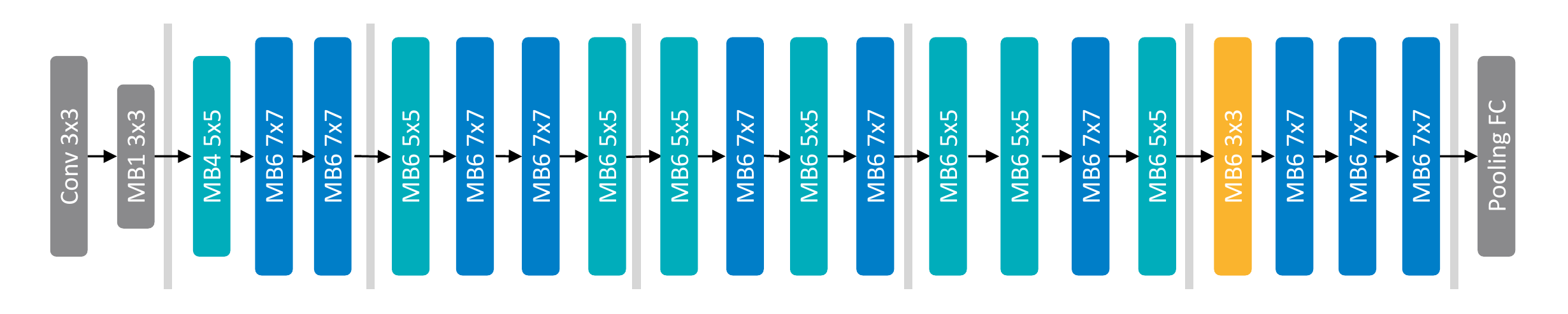}
\caption{
    The architecture searched by \sexynamenas in MobileNetV3-like search space.
}
\label{fig:nacnas_mb_arch}
\end{figure}

\begin{figure}[H]
\renewcommand\thefigure{C}
\centering
\subfigure[Normal cell.]
{
	\label{fig:normal}
	\includegraphics[width = 0.48\columnwidth]{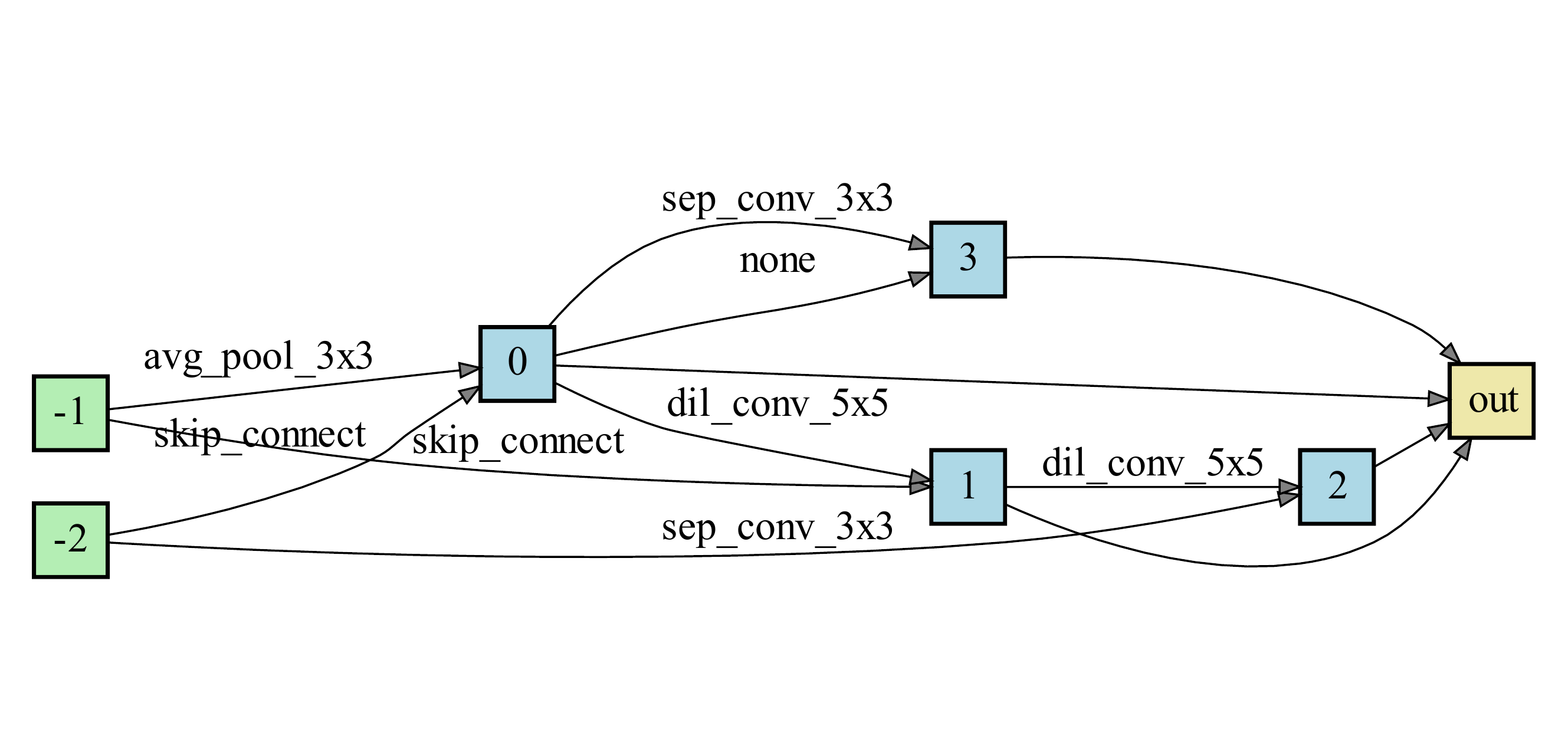}
}~~~
\subfigure[Reduction cell.]
{
	\label{fig:reduction}
	\includegraphics[width = 0.4\columnwidth]{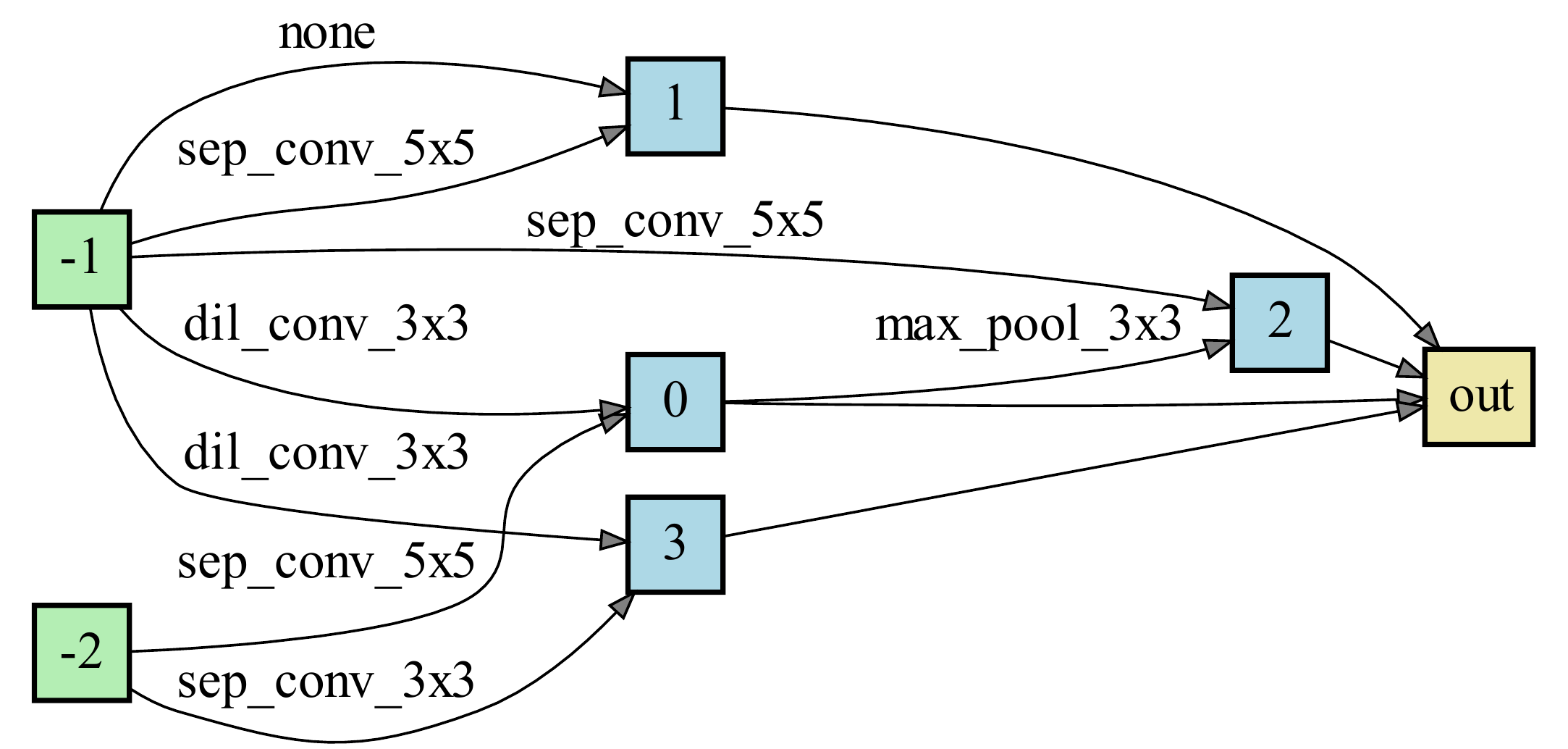}
}
\caption{The architecture of the convolutional cells found by \sexynamenas in DARTS search space.
\cyf{Following~\cite{liu2018darts}, we represent the convolutional cell as a directed acyclic graph, which consists of 7 nodes, including 2 input nodes (green boxes),  4 intermediate nodes (blue boxes) and 1 output node (yellow box). Each labeled edge in the cell denotes an operation (\eg, $3\times3$ separable convolution). The unlabeled edges denote the feature average operation over all intermediate nodes to obtain the output node.}
}
\label{fig:cnn_arch}
\end{figure}

% \clearpage
% {\small
% \bibliographystyle{ieee_fullname}
% \bibliography{egbib}
% }

}
\end{document}